\documentclass{CVM}
\usepackage{color}
\usepackage{xcolor,colortbl}
\usepackage{anyfontsize}

\newcommand{\wtmin}[1]{\textcolor{black}{#1}}

\newcommand{\longgs}{{3D Gaussian Splatting}\xspace}
\newcommand{\shortgs}{{3DGS}\xspace}

\newcommand{\bc}[1]{{\cellcolor[HTML]{ffb2b2}#1}}
\newcommand{\sbc}[1]{{\cellcolor[HTML]{ffcc99}#1}}
\newcommand{\tbc}[1]{{\cellcolor[HTML]{ffffb2}#1}}

\CVMsetup{
type      = {Research/Review Article},
doi       = {s41095-0xx-xxxx-x},
title     = {Recent Advances in 3D Gaussian Splatting},
author    = {Tong Wu$^{1}$, Yu-Jie Yuan$^{1}$, Ling-Xiao Zhang$^{1}$, Jie Yang$^{1}$, Yan-Pei Cao$^{2}$, Ling-Qi Yan$^{3}$, and Lin Gao$^{1}$\cor{}},
runauthor = {T. Wu, Y.-J Yuan, L.-X Zhang, J. Yang, Y.-P Cao, L.-Q Yan, L. Gao},
abstract  = {
  The emergence of \longgs (\shortgs) has greatly accelerated the rendering speed of novel view synthesis. 
Unlike neural implicit representations like Neural Radiance Fields (NeRF) that represent a 3D scene with position and viewpoint-conditioned neural networks, \longgs utilizes a set of Gaussian ellipsoids to model the scene so that efficient rendering can be accomplished by rasterizing Gaussian ellipsoids into images. 
Apart from the fast rendering speed, the explicit representation of \longgs facilitates downstream tasks like dynamic reconstruction, geometry editing, and physical simulation. 
Considering the rapid change and growing number of works in this field, we present a literature review of recent \longgs methods, which can be roughly classified into 3D reconstruction, 3D editing, and other downstream applications by functionality. 
Traditional point-based rendering methods and the rendering formulation of \longgs are also illustrated for a better understanding of this technique. 
This survey aims to help beginners get into this field quickly and provide 
experienced researchers with a comprehensive overview, which can stimulate the future development of the \longgs representation. 

},
keywords  = {\longgs, radiance field, novel view synthesis, editing, generation},
copyright = {The Author(s)},
}

\begin{document}

\maketitle

    \begin{figure}[b] \vskip -4mm
    \small\renewcommand\arraystretch{1.3}
        \begin{tabular}{p{80.5mm}} \toprule\\ \end{tabular}
        \vskip -4.5mm \noindent \setlength{\tabcolsep}{1pt}
        \begin{tabular}{p{3.5mm}p{80mm}}
$1\quad $ & Institute of Computing Technology, Chinese Academy of Sciences, Beijing, China, 100190. E-mail: Tong Wu, wutong19s@ict.ac.cn; Yu-Jie Yuan, yuanyujie@ict.ac.cn; Ling-Xiao Zhang, zhanglingxiao@ict.ac.cn; Jie Yang, yangjie01@ict.ac.cn; Lin Gao, gaolin@ict.ac.cn.\\
$2\quad $ & VAST. Yan-Pei Cao, caoyanpei@gmail.com.\\
$3\quad $ & Department of Computer Science, University of California. Ling-Qi Yan, lingqi@cs.ucsb.edu.\\

&\hspace{-5mm} Manuscript received: 2024-02-28;\vspace{-2mm}
    \end{tabular} \vspace {-3mm}
    \end{figure}

\section{Introduction}
\label{sec:introduction}
With the development of virtual reality and augmented reality, the demand for realistic 3D content is increasing. 
Traditional 3D content creation methods include 3D reconstruction from scanners or multi-view images and 3D modeling with professional software. 
However, traditional 3D reconstruction methods are likely to produce less faithful results due to imperfect capture and noisy camera estimation. 
3D modeling methods yield realistic 3D content but require professional user training and interaction, which is time-consuming. 

To create realistic 3D content automatically, Neural radiance fields (NeRF)~\cite{NeRF} propose to model a 3D scene's geometry and appearance with a density field and a color field, respectively. 
NeRF greatly improved the quality of novel view synthesis results but still suffered from its low training and rendering speed. 
While significant efforts~\cite{InstantNGP, MobileNeRF, BakedSDF} have been made to accelerate it to facilitate its applications on common devices, it is still hard to find a robust method that can both train a NeRF quickly enough ($\leq$ 1 hour)  on a consumer-level GPU and render a 3D scene at an interactive frame rate ($\sim$ 30 FPS) on common devices like cellphones and laptops. 
To resolve the speed issues, \longgs (\shortgs)~\cite{GS} proposes to rasterize a set of Gaussian ellipsoids to approximate the appearance of a 3D scene, which not only achieves comparable novel view synthesis quality but also allows fast converge ($\sim$ 30 minutes) and real-time rendering ($\geq$ 30 FPS) at 1080p resolution, making low-cost 3D content creation and real-time applications possible.

\begin{figure*}[!t]
\centering
\includegraphics[width=0.99\textwidth]{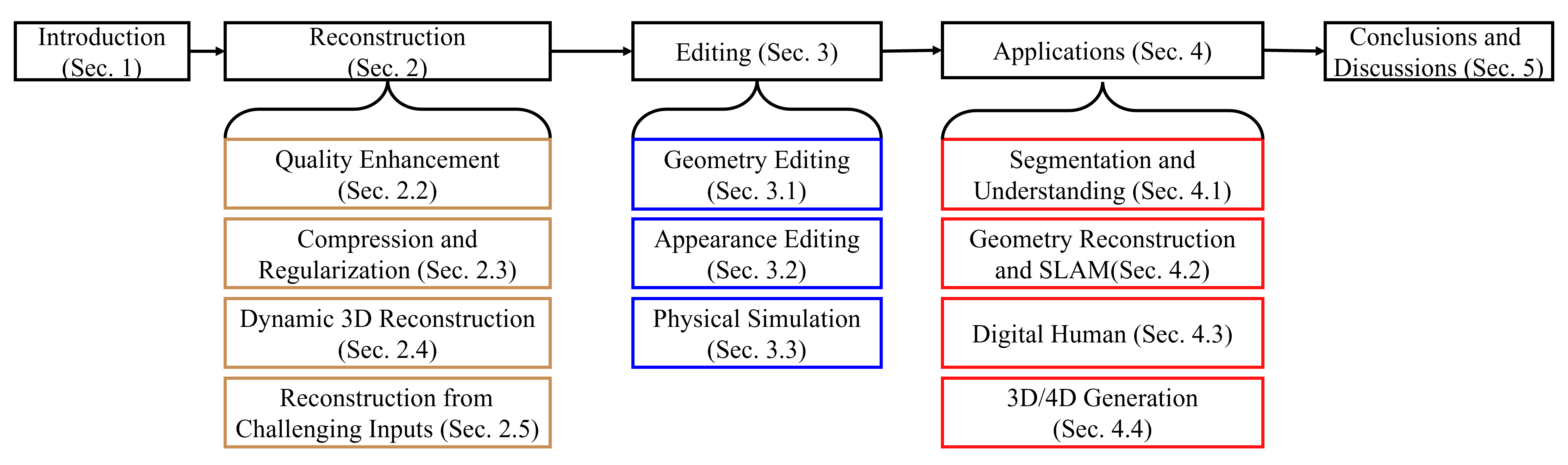}
    \vspace{-5pt}
\caption{Structure of the literature review and taxonomy of current \longgs methods.}
\label{fig:structure}
\vspace{-10pt}
\end{figure*}

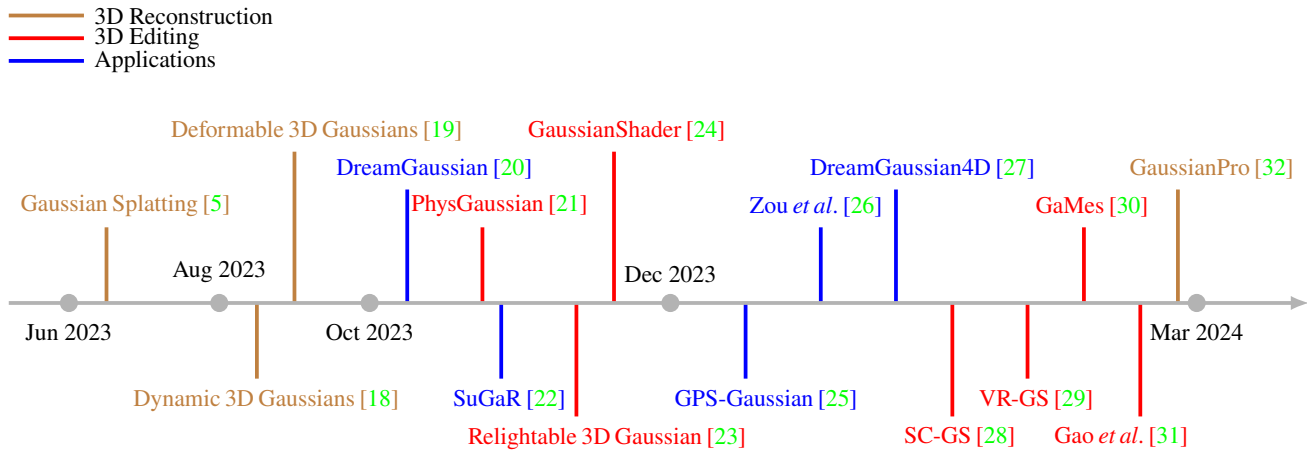
\begin{figure*}[!t]
\begin{minipage}{0.5\linewidth}
\begin{tikzpicture}[scale=1]

\tikzstyle{every node}=[font=\tiny,scale=1]
    \fill[gray!60] (1,4) circle (1.2mm) node[black, anchor=north, yshift=-1.5mm] {\small{Jun 2023}};
    \fill[gray!60] (3,4) circle (1.2mm) node[black, anchor=south, yshift=1.5mm] {\small{Aug 2023}};
    \fill[gray!60] (5,4) circle (1.2mm) node[black, anchor=north, yshift=-1.5mm] {\small{Oct 2023}};
    \fill[gray!60] (9,4) circle (1.2mm) node[black, anchor=south, yshift=1.5mm] {\small{Dec 2023}};
    \fill[gray!60] (16,4) circle (1.2mm) node[black, anchor=north, yshift=-1.5mm] {\small{Mar 2024}};

    \draw[brown, line width=0.5mm] (0.2, 7.8) -- (1.2, 7.8) node[black, anchor=west] {\small{3D Reconstruction}};
    \draw[red, line width=0.5mm] (0.2, 7.5) -- (1.2, 7.5) node[black, anchor=west] {\small{3D Editing}};
    \draw[blue, line width=0.5mm] (0.2, 7.2) -- (1.2, 7.2) node[black, anchor=west] {\small{Applications}};

    \draw[brown, line width=0.5mm] (1.5, 4) -- (1.5, 5) node[anchor=south, text width=3cm, xshift=3mm] {\small{ Gaussian Splatting~\cite{GS} }};
    \draw[brown, line width=0.5mm] (3.5, 4) -- (3.5, 3) node[anchor=north, text width=4cm, xshift=3mm] {\small{ Dynamic 3D Gaussians~\cite{luiten2023dynamic} }};
    \draw[brown, line width=0.5mm] (4.0, 4) -- (4.0, 6) node[anchor=south, text width=4cm, xshift=3mm] {\small{ Deformable 3D Gaussians~\cite{yang2023deformable} }};
    
    \draw[blue, line width=0.5mm] (5.5, 4) -- (5.5, 5.5) node[anchor=south, text width=4cm, xshift=10mm] {\small{ DreamGaussian~\cite{tang2023dreamgaussian} }};
    \draw[red, line width=0.5mm] (6.5, 4) -- (6.5, 5) node[anchor=south, text width=4cm, xshift=10mm] {\small{ PhysGaussian~\cite{xie2023physgaussian} }};
    \draw[blue, line width=0.5mm] (6.75, 4) -- (6.75, 3) node[anchor=north, text width=4cm, xshift=13mm] {\small{ SuGaR~\cite{SuGaR} }};
    \draw[red, line width=0.5mm] (7.75, 4) -- (7.75, 2.5) node[anchor=north, text width=4cm, xshift=5mm] {\small{ Relightable 3D Gaussian~\cite{RelightableGaussian} }};
    \draw[red, line width=0.5mm] (8.25, 4) -- (8.25, 6) node[anchor=south, text width=4cm, xshift=8mm] {\small{ GaussianShader~\cite{GaussianShader} }};
    
    \draw[blue, line width=0.5mm] (10, 4) -- (10, 3) node[anchor=north, text width=4cm, xshift=10mm] {\small{ GPS-Gaussian~\cite{zheng2023gps} }};
    \draw[blue, line width=0.5mm] (11, 4) -- (11, 5) node[anchor=south, text width=4cm, xshift=10mm] {\small{ Zou \etal~\cite{zou2023triplane} }};
    \draw[blue, line width=0.5mm] (12, 4) -- (12, 5.5) node[anchor=south, text width=4cm, xshift=8mm] {\small{ DreamGaussian4D~\cite{ren2023dreamgaussian4d} }};
    \draw[red, line width=0.5mm] (12.75, 4) -- (12.75, 2.5) node[anchor=north, text width=4cm, xshift=13mm] {\small{ SC-GS~\cite{huang2023sc} }};
    \draw[red, line width=0.5mm] (13.75, 4) -- (13.75, 3) node[anchor=north, text width=4cm, xshift=13mm] {\small{ VR-GS~\cite{jiang2024vr} }};
    \draw[red, line width=0.5mm] (14.5, 4) -- (14.5, 5) node[anchor=south, text width=3cm, xshift=8mm] {\small{ GaMes~\cite{waczynska2024games} }};
    \draw[red, line width=0.5mm] (15.25, 4) -- (15.25, 2.5) node[anchor=north, text width=4cm, xshift=8mm] {\small{ Gao \etal~\cite{gao2024mesh} }};
    \draw[brown, line width=0.5mm] (15.75, 4) -- (15.75, 5.5) node[anchor=south, text width=3cm, xshift=8mm] {\small{ GaussianPro~\cite{GaussianPro} }};

    \draw[gray!60, -latex, line width=0.5mm] (0.2, 4) -- (17.5, 4);
\end{tikzpicture}

\end{minipage}
\caption{{A brief timeline of representative works with the \longgs representation.} }
\label{fig:timeline}
\end{figure*}

Based on the \longgs representation, a number of research works have come out and more are on the way. 
To help readers get familiar with \longgs quickly, we present a survey on \longgs, which covers both traditional splatting methods and recent neural-based \shortgs methods. 
\wtmin{There are two literature reviews~\cite{survey_chen,survey_fei} that summarize recent works on \shortgs, which also serve as a good reference tool for this field.}
As shown in Fig.~\ref{fig:structure}, we divide these works into three parts by functionality. 
We first introduce how \longgs allows realistic scene reconstruction under various sceneries (Sec.~\ref{sec:recon}) and further present scene editing techniques with \longgs (Sec.~\ref{sec:edit}) and how \longgs makes downstream applications like digital human possible (Sec.~\ref{sec:applications}). 
Finally, we summarize recent research works on \longgs at a higher level and present future works remaining to be done in this field (Sec.~\ref{sec:conclusion}). 
A timeline of representative works can be found in Fig.~\ref{fig:timeline}.

\section{Gaussian Splatting for 3D Reconstruction}
\label{sec:recon}
\subsection{Point-based Rendering}
Point-based rendering technique aims to generate realistic images by rendering a set of discrete geometry primitives.
Grossman and Dally~\cite{psr} propose the point-based rendering technique based on the purely point-based representation, where each point only influences one pixel on the screen.
Instead of rendering points, Zwicker \etal ~\cite{surface_splatting} propose to render splats (ellipsoids) so that each splat can occupy multiple pixels and the mutual overlap between them can generate hole-free images more easily than purely point-based representation.
Later, a series of splatting methods aim to enhance it by introducing a texture filter for anti-aliasing rendering~\cite{EWA_splatting}, improving rendering efficiency~\cite{BotschWK02, BotschK03}, and resolving discontinuous shading~\cite{RusinkiewiczL00}.
For more details about traditional point-based rendering techniques, please refer to~\cite{survey_leif}.

Traditional point-based rendering methods focus more on how to produce high-quality rendered results with a given geometry.
With the development of recent implicit representation~\cite{IMNet, DeepSDF, OccNet}, researchers have started to explore point-based rendering with the neural implicit representation without any given geometry for the 3D reconstruction task.
One representative work is NeRF~\cite{NeRF} which models geometry with an implicit density field and predicts view-dependent color $c_i$ with another appearance field.
The point-based rendering combines all sample points' color on a camera ray to produce a pixel color $C$ by:
\begin{equation}
\label{eqn:pbr}
C = \sum_{i=1}^N c_i \alpha_i T_i
\end{equation}
where $N$ is the number of sample points on a ray and $\alpha_i = exp(-\sum_{j=1}^{i-1} \sigma_j \delta_j)$ are the view-dependent color and the opacity value for $i$th point on the ray.
$\sigma_j$ is the $j$th point's density value.
$T_i = \prod_{j=1}^{i-1} (1-\alpha_j)$ is accumulated transmittance.
To ensure high-quality rendering, NeRF~\cite{NeRF} typically requires sampling 128 points on a single ray, which unavoidably takes longer time to train and render.

To speed up both training and rendering, instead of predicting density values and colors for all sample points with neural networks, \longgs~\cite{GS} abandons the neural network and directly optimizes Gaussian ellipsoids to which attributes like position $P$, rotation $R$, scale $S$, opacity $\alpha$ and Spherical Harmonic coefficients (SH) representing view-dependent color are attached.
The pixel color is determined by Gaussian ellipsoids projected onto it from a given viewpoint.
The projection of 3D Gaussian ellipsoids can be formulated as:
\begin{equation}
    \Sigma' = JW\Sigma W^T J^T
\end{equation}
\wtmin{where $\Sigma'$ and $\Sigma=RSS^T R^T$ are the covariance matrices for 3D Gaussian ellipsoids and projected Gaussian ellipsoids on 2D image from a viewpoint with viewing transformation matrix $W$.}
$J$ is the Jacobian matrix for the projective transformation.
\shortgs shares a similar rendering process with NeRF but there are two major differences between them:
\begin{enumerate}
    \item \shortgs models opacity values directly while NeRF transforms density values to the opacity values.
    \item \shortgs uses rasterization-based rendering which does not require sampling points while NeRF requires dense sampling in the 3D space.
\end{enumerate}
Without sampling points and querying the neural network, \shortgs becomes extremely fast and achieves $\sim$ 30 FPS on a common device with comparable rendering quality with NeRF.

\subsection{Quality Enhancement}
Though producing high-quality reconstruction results, there are still improvement spaces for \shortgs's rendering.
Mip-Splatting~\cite{Mip-Splatting} observed that changing the sampling rate, for example, the focal length, can greatly influence the quality of rendered images by introducing high-frequency Gaussian shape-like artifacts or strong dilation effects.
To eliminate the high-frequency Gaussian shape-like artifacts, Mip-Splatting~\cite{Mip-Splatting} constrains the frequency of the 3D representation to be below half the maximum sampling frequency determined by the training images.
In addition, to avoid the dilation effects, it introduces another 2D Mip filter to projected Gaussian ellipsoids to approximate the box filter similar to EWA-Splatting~\cite{EWA_splatting}.
MS3DGS~\cite{MS3DGS} also aims at solving the aliasing problem in the original \shortgs and introduces a multi-scale Gaussian splatting representation and when rendering a scene at a novel resolution level, it selects Gaussians from different scale levels to produce alias-free images.
\wtmin{Analytic-Splatting~\cite{Analytic-Splatting} approximates the cumulative distribution function of Gaussians with a logistic function to better model each pixel's intensity response for anti-alias.
SA-GS~\cite{SA-GS} utilizes an adaptive 2D low-pass filter at the test time according to the rendering resolution and camera distance.}

\begin{figure*}[!t]
    \centering
    \includegraphics[width=0.99\linewidth]{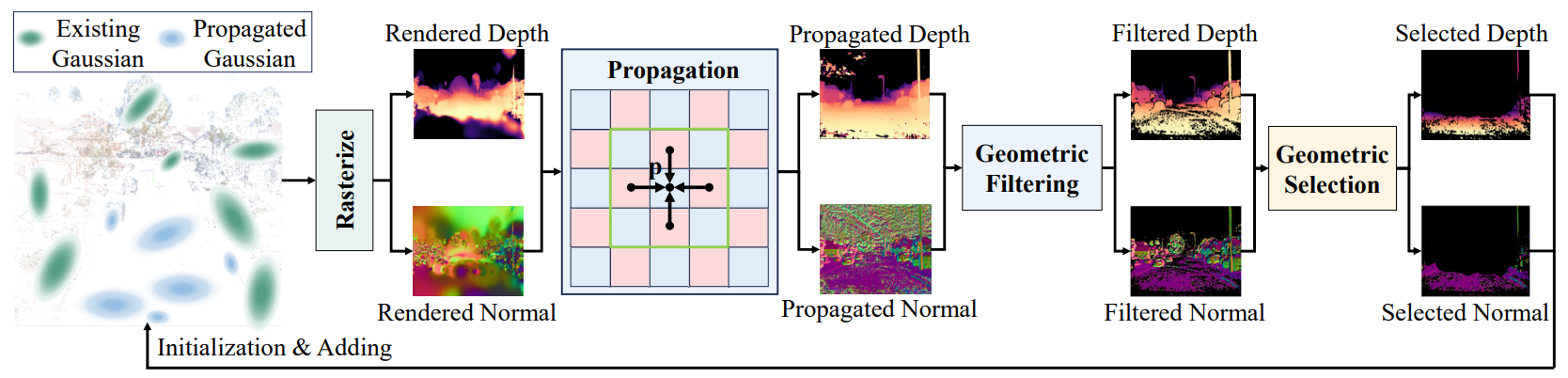}
    \caption{Overview of GaussianPro~\cite{GaussianPro}. Neighboring views' normal direction consistency is considered to produce better reconstruction results.}
    \label{fig:recon}
\end{figure*}

Apart from the aliasing problem, the capability of rendering view-dependent effects also needs to be improved.
To produce more faithful view-dependent effects, VDGS~\cite{VDGS} proposes to model the \shortgs to represent 3D shapes and predict attributes like view-dependent color and opacity with NeRF-like neural network instead of Spherical Harmonic (SH) coefficients in the original \shortgs.
Scaffold-GS~\cite{Scaffold-GS} proposes to initialize a voxel grid and attach learnable features onto each voxel point and all attributes of Gaussians are determined by interpolated features and lightweight neural networks. 
\wtmin{Based on Scaffold-GS, Octree-GS~\cite{Octree-GS} introduces a level-of-detail strategy to better capture details.}
Instead of changing the view-dependent appearance modeling approach, StopThePop~\cite{StopThePop} points out that \shortgs tends to cheat view-dependent effects by popping 3D Gaussians due to the per-ray depth sorting, which leads to less faithful results when the viewpoint is rotated.
To mitigate the potential of popping 3D Gaussians, StopThePop~\cite{StopThePop} replaces the per-ray depth sorting with tile-based sorting to ensure consistent sorting order at a local region.
To better guide the growth of \longgs, GaussianPro~\cite{GaussianPro} introduces a progressive propagation strategy to updating Gaussians by considering the normal consistency between neighboring views and adding plane constraints as shown in Fig.~\ref{fig:recon}.
\wtmin{GeoGaussian~\cite{GeoGaussian} proposes to densify Gaussians on the tangent plane of Gaussians and encourage the smoothness of geometric properties between neighboring Gaussians.
RadSplat~\cite{RadSplat} initializes \shortgs with point cloud derived from a trained neural radiance field and prunes Gaussians with a multi-view importance score.}
To deal with more complex shadings like specular and anisotropic components, Spec-Gaussian~\cite{Spec-Gaussian} proposes to utilize Anisotropic Spherical Gaussian to approximate 3D scenes' appearance.
\wtmin{TRIPS~\cite{TRIPS} attaches a neural feature onto Gaussians and renders pyramid-like image feature planes according to the projected Gaussians size similar to ADOP~\cite{ADOP} to resolve the blur issue in the original \shortgs.
Handling the same issue, FreGS~\cite{FreGS} applies frequency-domain regularization to the rendered 2D image to encourage high-frequency details recovery.
GES~\cite{GES} utilizes the Generalized Normal Distribution(NFD) to sharpen scene edges. 
To resolve the problem that \shortgs is sensitive to initialization, RAIN-GS~\cite{RAIN-GS} proposes to initialize Gaussians sparsely with large variance from the SfM point cloud and progressively applies low-pass filtering to avoid 2D Gaussian projections smaller than a pixel.
Pixel-GS~\cite{Pixel-GS} takes the number of pixels that a Gaussian covers from all the input viewpoints into account in the splitting process and scales the gradient according to the distance to the camera to suppress floaters.
Bul\`{o}~\etal~\cite{RevisingDensification} also leverage pixel-level error as a densification criteria and revise the opacity setting in the cloning process for more stable training process.
}
Quantitative results of different reconstruction methods can be found in Table~\ref{tab:nvs}.
\shortgs-based methods and NeRF-based methods are comparable but \shortgs-based methods have faster rendering speed.

\begin{table}[t!]
    \caption{
        Quantitative comparison of novel view synthesis results on the MipNeRF 360 dataset~\cite{barron2022mipnerf360} using PSNR, SSIM and LPIPS metrics.
    }
        \centering
                \setlength\tabcolsep{0pt}
                \begin{tabular*}{\columnwidth}{@{\extracolsep{\fill}} cccc}
\hline
                    Methods                                 & PSNR $\uparrow$ & SSIM $\uparrow$ & LPIPS $\downarrow$ \\ \hline
                    MipNeRF~\cite{barron2021mip}            & 24.04           & 0.616           & 0.441              \\
                    MipNeRF 360~\cite{barron2022mipnerf360} & 27.57           & 0.793           & 0.234              \\
                    ZipNeRF~\cite{barron2023zipnerf}        & \sbc{28.54}     & \sbc{0.828}     & \bc{0.189}         \\ \hline
                    \shortgs~\cite{GS}                      & 27.21           & 0.815           & 0.214              \\
                    Mip-Splatting~\cite{Mip-Splatting}      & 27.79           & \tbc{0.827}     & \sbc{0.203}        \\
                    Scaffold-GS~\cite{Scaffold-GS}          & \bc{28.84}      & \bc{0.848}      & 0.220              \\
                    VDGS~\cite{VDGS}                        & 27.66           & 0.809           & 0.223              \\
                    GaussianPro~\cite{GaussianPro}          & \tbc{27.92}     & 0.825           & \tbc{0.208}        \\ \hline
                \end{tabular*}
    \label{tab:nvs}
\end{table}

\subsection{Compression and Regularization}
Although the \longgs achieves real-time rendering, there is improvement space in terms of lower computational requirements and better point distribution.
Some methods focus on changing the original representation to reduce computational resources.

\begin{figure*}[!t]
    \centering
    \includegraphics[width=0.99\linewidth]{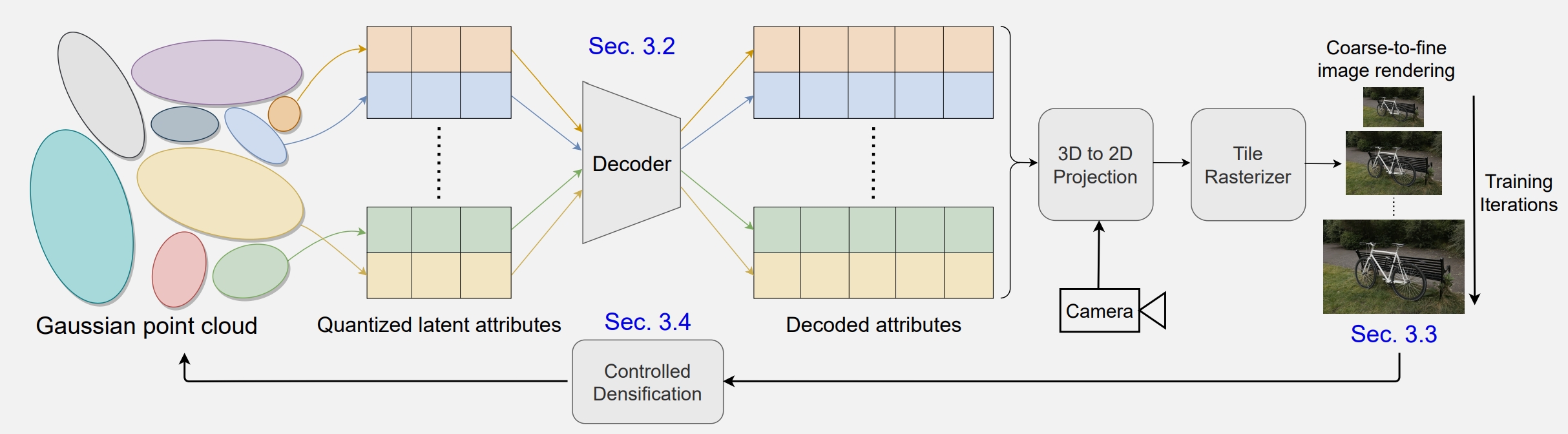}
    \caption{Pipeline from EAGLES~\cite{girish2023eagles}. Vector Quantization (VQ) is utilized to compress Gaussian attributes.}
    \label{fig:compression}
\end{figure*}

Vector Quantization, a traditional compression method in signal processing, which involves clustering multi-dimensional data into a finite set of representations, is mainly utilized in Gaussians~\cite{lee2023compact,navaneet2023compact3d,niedermayr2023compressed,girish2023eagles,fan2023lightgaussian}.
C3DGS~\cite{lee2023compact} adopts residual vector quantization (R-VQ)~\cite{zeghidour2021soundstream} to represent geometric attributes, including scaling and rotation.
SASCGS~\cite{niedermayr2023compressed} utilizes vector clustering to encode color and geometric attributes into two codebooks, with a sensitivity-aware K-Means method.
As shown in Fig.~\ref{fig:compression}, EAGLES~\cite{girish2023eagles} quantizes all attributes including color, position, opacity, rotation, and scaling, they show that the quantization of opacity leads to fewer floaters or visual artifacts in the novel view synthesis task.
Compact3D~\cite{navaneet2023compact3d} does not quantize opacity and position, because sharing them results in overlapping Gaussians.
LightGaussian~\cite{fan2023lightgaussian} prunes Gaussians with a smaller importance score and adopts octree-based lossless compression in G-PCC~\cite{pcc} for the position attribute due to the sensitivity to the subsequent rasterization accuracy for the position.
\wtmin{Based on the same importance score calculation, Mini-Splatting~\cite{Mini-Splatting} samples Gaussians instead of pruning points to avoid artifacts caused by pruning.}
SOGS~\cite{morgenstern2023compact} adopt a different method from Vector Quantization.
They arrange Gaussian attributes into multiple 2D grids.
These grids are sorted and a smoothness regularization is applied to penalize all pixels that have very different values compared to their local neighborhood on the 2D grid.
\wtmin{HAC~\cite{HAC} adopts the idea of Scaffold-GS~\cite{Scaffold-GS} to model the scene with a set of anchor points and learnable features on these anchor points. 
It future introduces an adaptive quantization module to compress the features of anchor points with the multi-resolution hash grid~\cite{InstantNGP}. 
Jo~\etal~\cite{jo2024identifying} propose to identify unnecessary Gaussians to compress \shortgs and accelerate the computation. 
Apart from 3D compression, the \longgs has also been applied to 2D image compression~\cite{GaussianImage}, where the 3D Gaussians degenerates to 2D Gaussians.}

In terms of disk data storage, SASCGS~\cite{niedermayr2023compressed} utilizes the entropy encoding method DEFLATE, which utilizes a combination of the LZ77 algorithm and Huffman coding, to compress the data.
SOGS~\cite{morgenstern2023compact} compress the RGB grid with JPEG XL and store all other attributes as 32-bit OpenEXR images with zip compression.
Quantitative reconstruction results and sizes of 3D scenes after compression are shown in Table~\ref{tab:compression}.

\begin{table}[!t]
    \center
    \caption{Comparison of different compression methods on the MipNeRF360~\cite{barron2022mipnerf360} dataset. Size is measured in MB. }
    \begin{tabular}{l|>{\centering\arraybackslash}m{8mm}>{\centering\arraybackslash}m{8mm}>{\centering\arraybackslash}m{8mm}>{\centering\arraybackslash}m{6mm}}
    \toprule
    Methods &  SSIM$\uparrow$   & PSNR$\uparrow$    & LPIPS$\downarrow$ & Size$\downarrow$\\
    \midrule
    3DGS\cite{GS}& \sbc{0.815} & \sbc{27.21} &  \sbc{0.214}  & 750\\
    C3DGS\cite{lee2023compact}& 0.798 & 27.08 & 0.247  & 48.8   \\
    Compact3D~\cite{navaneet2023compact3d}& 0.808 & \tbc{27.16} & \tbc{0.228}  & /  \\
    EAGLES~\cite{girish2023eagles}& \tbc{0.81} & 27.15 &  0.24  &  68 \\
    SOGS\cite{morgenstern2023compact}& 0.763 & 25.83 & 0.273  & \bc{18.2}  \\
    SASCGS\cite{niedermayr2023compressed}& 0.801 & 26.98 & 0.238  & \tbc{28.80} \\
    LightGaussian~\cite{fan2023lightgaussian}& \bc{0.857} & \bc{28.45} & \bc{0.210} & \sbc{42.48} \\
    \bottomrule
    \end{tabular}
    \label{tab:compression}
    \end{table}

\subsection{Dynamic 3D Reconstruction}
\begin{figure*}[!ht]
    \centering
    \includegraphics[width=0.99\textwidth]{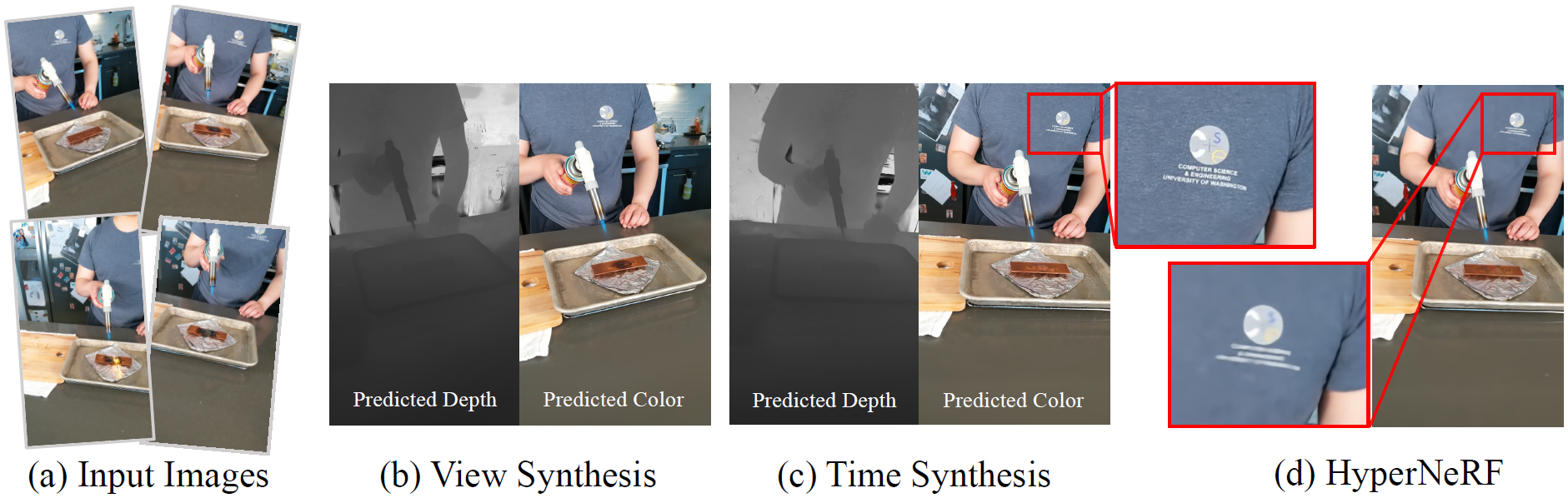}
    \caption{The results of Deformable3DGS~\cite{yang2023deformable}. Given a set of monocular multi-view images (a), this method can achieve novel view synthesis (b) and time synthesis (c), and has better rendering quality compared to HyperNeRF~\cite{park2021hypernerf} (d).}
    \label{fig:dynamic}
\end{figure*}

The same as the NeRF representation, \shortgs can also be extended to reconstruct dynamic scenes.
The core of dynamic \shortgs lies in how to model the variations of Gaussian attribute values over time.
The most straightforward way is to assign different attribute values to 3D Gaussian at different timesteps.
Luiten \etal~\cite{luiten2023dynamic} regard the center and rotation (quaternion) of 3D Gaussian as variables that change over time, while other attributes remain constant over all timesteps, thus achieving 6-DOF tracking by reconstructing dynamic scenes.
However, the frame-by-frame discrete definition lacks continuity, which can cause poor results in long-term tracking.
Therefore, physical-based constraints are introduced, which are three regularization losses, including short-term local-rigidity and local-rotation similarity losses and a long-term local-isometry loss.
However, this method still lacks inter-frame correlation and requires high storage overhead for long-term sequences.
Therefore, decomposing spatial and temporal information and modeling them with a canonical space and a deformation field, respectively, has become another exploration direction.
The canonical space is the static \shortgs, then the problem becomes how to model the deformation field.
One way is to use an MLP network to implicitly fit it, similar to the dynamic NeRF~\cite{pumarola2021d}.
Yang \etal~\cite{yang2023deformable} follow this idea and propose to input the positional-encoded Gaussian position and timestep $t$ to the MLP which outputs the offsets of the position, rotation, and scaling of 3D Gaussian.
However, inaccurate poses may affect rendering quality.
This is not significant in continuous modeling of NeRF, but discrete \shortgs can amplify this problem, especially in the time interpolation task.
So, they add a linearly decaying Gaussian noise to the encoded time vector to improve temporal smoothing without additional computational overhead. Some results are shown in Fig.~\ref{fig:dynamic}.
4D-GS~\cite{wu20234d} adopts the multi-resolution HexPlane voxels~\cite{cao2023hexplane} to encode the temporal and spatial information of each 3D Gaussian rather than positional encoding and utilizes different compact MLPs for different attributes.
For stable training, it first optimizes static \shortgs and then optimizes the deformation field represented by an MLP.
GauFRe~\cite{liang2023gaufre} applies the exponential and normalization operation to the scaling and rotation respectively after adding the delta values predicted by an MLP, ensuring convenient and reasonable optimization.
As dynamic scenes contain large static parts, it randomly initializes the point cloud into dynamic point clouds and static point clouds, optimizes them accordingly, and renders them together to achieve decoupling of the dynamic part and the static part.
\wtmin{3DGStream~\cite{3DGStream} allows online training of \shortgs into the dynamic scene reconstruction by modeling the transformation between frames as a neural transformation cache and adaptively adding 3D Gaussians to handle emerging objects.
4DGaussianSplatting~\cite{4DGaussianSplatting} turns 3D Gaussians into 4D Gaussians and slice the 4D Gaussians into 3D Gaussians for each time step. 
The sliced 3D Gaussians are projected onto image plane to reconstruct the corresponding frame.}
\wtmin{Guo~\etal~\cite{guo2024motionaware} and GaussianFlow~\cite{GaussianFlow} introduce 2D flow estimation results into the training of dynamic 3D Gaussians, which supports deformation modeling between neighboring frames and enables superior 4D reconstruction and 4D generation results.
TOGS~\cite{TOGS} constructs an opacity offset table to model the changes in digital subtraction angiography.
Zhang~\etal~\cite{zhang2024bags} leverage the diffusion prior to enhance dynamic scene reconstruction and propose a neural bone transformation module for animatable objects reconstruction from monocular videos.}

Compared to NeRF, \shortgs is an explicit representation and the implicit deformation modeling requires lots of parameters which may bring overfit, so some explicit deformation modeling methods are also proposed, which ensure fast training.
Katsumata \etal~\cite{katsumata2023efficient} propose to use the Fourier series to fit the changes of the Gaussian position, inspired by the fact that the motion of human and articulated objects is sometimes periodic. The rotation is approximated by a linear function. Other attributes remain unchanged over time. So dynamic optimization is to optimize the parameters of the Fourier series and the linear function, and the number of parameters is independent of time. These parametric functions are continuous functions about time, ensuring temporal continuity and thus ensuring the robustness of novel view synthesis. In addition to the image losses, a bidirectional optical flow loss is also introduced.
The polynomial fitting and Fourier approximation have advantages in modeling smooth motion and violent motion, respectively. So Gaussian-Flow~\cite{lin2023gaussian} combines these two methods in the time and frequency domains to capture the time-dependent residuals of the attribute, named as Dual-Domain Deformation Model (DDDM). The position, rotation, and color are considered to change over time. To prevent optimization problems caused by uniform time division, this work adopts adaptive timestep scaling.
Finally, the optimization iterates between static optimization and dynamic optimization, and introduces temporal smoothness loss and KNN rigid loss.
Li \etal~\cite{li2023spacetime} introduce a temporal radial basis function to represent temporal opacity, which can effectively model the scene content that emerges or vanishes. 
Then, the polynomial function is exploited to model the motion and rotation of 3D Gaussians. 
They also replace the spherical harmonics with features to represent view- and time-related color. 
These features consist of three parts: base color, view-related feature, and time-related feature. 
The latter two are translated into a residual color through an MLP added to the base color to obtain the final color. 
During optimization, new 3D Gaussians will be sampled at the under-optimized positions based on training error and coarse depth.
The explicit modeling methods used in the above methods are all based on commonly used functions.
DynMF~\cite{kratimenos2023dynmf} assumes that each dynamic scene is composed of a finite and fixed number of motion trajectories and argues that a learned basis of the trajectories will be smoother and more expressive. 
All motion trajectories in the scene can be linearly represented by this learned basis and a small temporal MLP is used to generate the basis. 
The position and rotation change over time and both share the motion coefficients with different motion bases. 
The regularization, sparsity, and local rigidity terms of the motion coefficients are introduced during optimization.

\begin{table}[!t]
    \caption{
        Quantitative comparison of novel view synthesis results on the D-NeRF~\cite{pumarola2021d} dataset using PSNR, SSIM and LPIPS metrics.
    }
        \centering
                \setlength\tabcolsep{0pt}
                \begin{tabular*}{\columnwidth}{@{\extracolsep{\fill}} cccc}
\hline
                    Methods                              & PSNR $\uparrow$ & SSIM $\uparrow$    & LPIPS $\downarrow$ \\ \hline
                    D-NeRF~\cite{pumarola2021d}          & 31.69           & 0.975              & 0.057              \\
                    TiNeuVox~\cite{TiNeuVox}             & 33.76           & 0.984              & 0.044              \\
                    Tensor4D~\cite{Tensor4D}             & 27.72           & 0.945              & 0.051              \\
                    K-Planes~\cite{kplanes}              & 32.32           & 0.973              & 0.038              \\ \hline
                    CoGS~\cite{yu2023cogs}               & \tbc{37.90}     & 0.983              & 0.027              \\
                    GauFRe~\cite{liang2023gaufre}        & 34.80           & 0.985              & 0.028              \\
                    4D-GS~\cite{wu20234d}                & 34.01           & \tbc{0.989}        & \tbc{0.025}        \\
                    Yang \etal~\cite{yang2023deformable} & \sbc{39.51}     & \sbc{0.990}        & \sbc{0.012}        \\
                    SC-GS~\cite{huang2023sc}             & \bc{43.30}      & \bc{0.997}         & \bc{0.0078}        \\ \hline
                \end{tabular*}
    \label{tab:dynamic_nvs}
\end{table}

There are also some other ways to explore.
4DGS~\cite{yang2023real} regards the spacetime of the scene as an entirety and transforms 3D Gaussian into 4D Gaussian, that is, transforming the attribute values defined on Gaussian to 4D space. 
For example, the scaling matrix is diagonal, so adding a scaling factor of time dimension on the diagonal forms the scaling matrix in 4D space. The 4D extension of the spherical harmonics (SH) can be expressed as the combination of SH with 1D-basis functions.
SWAGS~\cite{shaw2023swags} divides the dynamic sequence into different windows based on the amount of motion and trains separate dynamic \shortgs model in different windows, with different canonical spaces and deformation fields. 
The deformation field uses a tunable MLP~\cite{maggioni2023tunable}, which focuses more on modeling the dynamic part of the scene. 
Finally, fine-tuning ensures temporal consistency between windows using the overlapping frame to add constraints. 
The MLP is fixed and only the canonical representation is optimized during fine-tuning.

These dynamic modeling methods can be further applied in the medical field, such as the markless motion reconstruction for motion analysis of infants and neonates~\cite{cotton2024dynamic} which introduces additional mask and depth supervisions, and monocular endoscopic reconstruction~\cite{EndoGS,chen2024endogaussians,huang2024endo,EndoGSLAM}.
Quantitative reconstruction results by representative NeRF-based and \shortgs-based methods are reported in Table~\ref{tab:dynamic_nvs}.
\wtmin{\shortgs-based methods have clear advantages compared to NeRF-based methods due to their explicit geometry representation that can model the dynamics more easily.
The efficient rendering of \shortgs also avoids densely sampling and querying the neural fields in NeRF-based methods and makes downstream applications of dynamic reconstruction like free-viewpoint video more feasible.}

\subsection{3D Reconstruction from Challenging Inputs}
\begin{figure*}[h]
    \centering
    \includegraphics[width=0.99\linewidth]{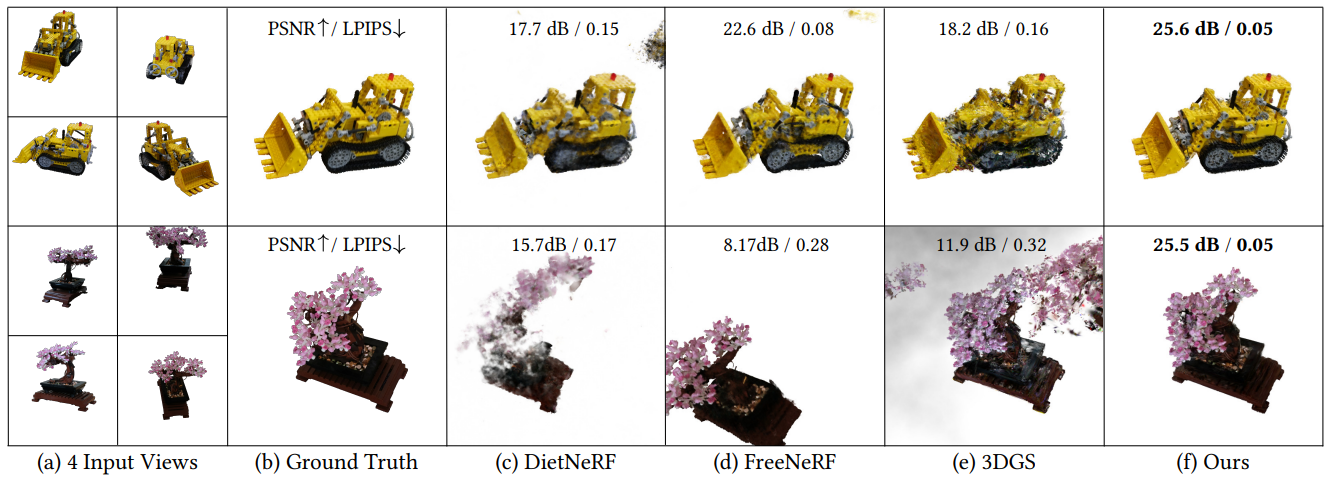}
    \caption{Results by GaussianObject~\cite{GaussainObject}. Compared to previous NeRF-based sparse-view reconstruction methods~\cite{DietNeRF, FreeNeRF} and recent \shortgs~\cite{GS}, GaussianObject achieves high-quality 3D reconstruction from only 4 views. }
    \label{fig:recon_challenging_input}
\end{figure*}

While most methods experiment on regular input data with dense viewpoints in relatively small scenes, there are also works targeting reconstructing 3D scenes with challenging inputs like sparse-view input, data without camera parameters, and larger scenes like urban streets.
FSGS~\cite{FSGS} is the first to explore reconstructing 3D scenes from sparse view input.
It initializes sparse Gaussians from structure-from-motion (SfM) methods and identifies them by unpooling existing Gaussians.
To allow faithful geometry reconstruction, an extra pre-trained 2D depth estimation network helps to supervise the rendered depth images.
SparseGS~\cite{SparseGS}, CoherentGS~\cite{CoherentGS}, and DNGaussian~\cite{DNGaussian} also target 3D reconstruction from sparse-view inputs by introducing depth inputs estimated by pre-trained a 2D network.
It further removes Gaussians with incorrect depth values and utilizes the Score Distillation Sampling (SDS) loss~\cite{poole2022dreamfusion} to encourage rendered results from novel viewpoints to be more faithful.
GaussainObject~\cite{GaussainObject} instead initializes Gaussians with visual hull and fine-tunes a pre-trained ControlNet~\cite{ControlNet} repair degraded rendered images generated by adding noises to Gaussians' attributes, which outperforms previous NeRF-based sparse-view reconstruction methods as shown in Fig.~\ref{fig:recon_challenging_input}.
Moving a step forward, pixelSplat~\cite{pixelSplat} reconstructs 3D scenes from single-view input without any data priors.
It extracts pixel-aligned image features similar to PixelNeRF~\cite{pixelNeRF} and predicts attributes for each Gaussian with neural networks.
MVSplat~\cite{MVSplat} brings the cost volume representation into sparse view reconstruction, which is taken as input for the attributes prediction network for Gaussians.
SplatterImage~\cite{SplatterImage} also works on single-view data but instead utilizes a U-Net~\cite{U-Net} network to translate the input image into attributes on Gaussians.
It can extend to multi-view inputs by aggregating predicted Gaussians from different viewpoints via warping operation.

For urban scene data, PVG~\cite{PeriodicVibrationGaussian} makes Gaussian's mean and opacity value time-dependent functions centered at corresponding Gaussian's life peak (maximum prominence over time).
DrivingGaussian~\cite{DrivingGaussian} and HUGS~\cite{HUGS} reconstruct dynamic driving data by first incrementally optimizing static 3D Gaussians and then composing them with dynamic objects' 3D Gaussians.
This process is also assisted by the Segmentation Anything Model~\cite{SAM} and input LiDAR depth data.
StreetGaussians~\cite{StreetGaussians} models the static background with a static \shortgs and dynamic objects by a dynamic \shortgs where Gaussians are transformed by tracked vehicle poses and their appearance is approximated with time-related Spherical Harmonics (SH) coefficients. 
SGD~\cite{SGD} incorporate diffusion priors with street scene reconstruction to improve the novel view synthesis results similar to ReconFusion~\cite{ReconFusion}.
HGS-Mapping~\cite{HGS-Mapping} separately models textureless sky, ground plane, and other objects for more faithful reconstruction. 
\wtmin{VastGaussian~\cite{VastGaussian} divides a large scene into multiple regions based on the camera distribution projected on the ground and learns to reconstruct a scene by iteratively adding more viewpoints into training based on visibility criteria. 
In addition, it models the appearance changes with an optimizable appearance embedding for each view.
CityGaussian~\cite{CityGaussian} also models the large-scale scene with a divide-and-conquer strategy and further introduces level-of-detail rendering based on the distance between the camera to a Gaussian.}
To facilitate comparisons on the urban scenes for \shortgs methods, GauU-Scene~\cite{GauU-Scene} provides a large-scale dataset covering over $1.5 km^2$.

Apart from the works mentioned above, other methods focus on special input data including images without camera~\cite{COLMAPFreeGS,iComMa,InstantSplat,GGRt}, blurry inputs~\cite{Deblur3DGS,BAGS,BAD-Gaussians,seiskari2024gaussian}, unconstrained images~\cite{SWAG,GaussianintheWild}, mirror-like inputs~\cite{Mirror-3DGS,comi2024snapit}, CT scans~\cite{SparseCT,X-Gaussian}, panoramic images~\cite{360-GS}, and satellite images~\cite{SatelliteGS}.

\section{Gaussian Splatting for 3D Editing}
\label{sec:edit}

\shortgs allows for efficient training and high-quality real-time rendering using rasterization-based point-based rendering techniques. 
Editing in \shortgs has been investigated in a number of fields.
We have summarized the editing on \shortgs into three categories: geometry editing, appearance editing, and physical simulation.

\subsection{Geometry Editing}
\begin{figure*}[!t]
    \centering
    \includegraphics[width=0.99\linewidth]{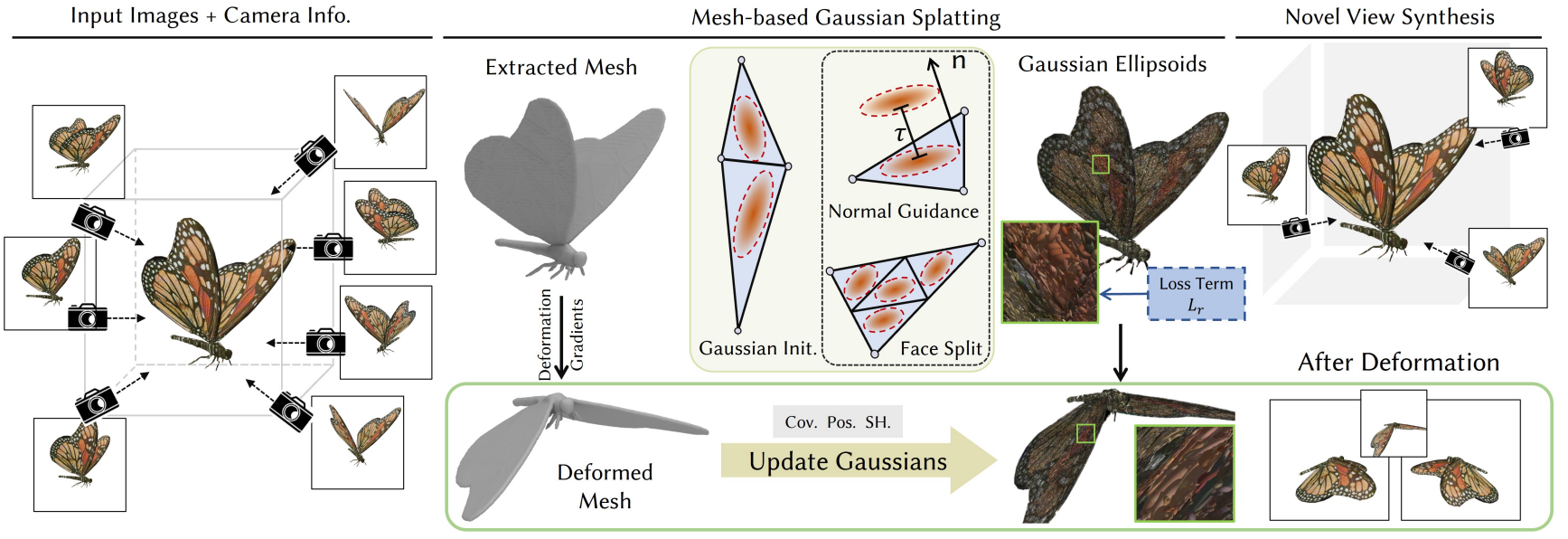}
    \caption{Pipeline of Gao \etal~\cite{gao2024mesh}. It allows large-scale geometry editing by binding 3D Gaussians onto the mesh.}
    \label{fig:edit_geo}
\end{figure*}

On the geometry side, 
GaussianEditor~\cite{GaussianEditor} controls the \shortgs using the text prompts and semantic information from proposed Gaussian semantic tracing, which enables 3D inpainting, object removal, and object composition.
Gaussian Grouping~\cite{GaussianGrouping} simultaneously rebuilds and segments open-world 3D objects under the supervision of 2D mask predictions from SAM and 3D spatial consistency constraints, which further enables diverse editing applications including 3D object removal, inpainting, and composition with high-quality visual effects and time efficiency.
Furthermore, Point'n Move~\cite{huang2023pointn} combines interactive scene object manipulation with exposed region inpainting. 
Thanks to the explicit representation of \shortgs, the dual-stage self-prompting mask propagation process is proposed to transfer the given 2D prompt points to 3D mask segmentation, resulting in a user-friendly editing experience with high-quality effects. 
\wtmin{Feng~\etal~\cite{feng2024new} propose a new Gaussian splitting algorithm to avoid inhomogeneous 3D Gaussian reconstruction and makes the boundary of 3D scenes after removal operation sharper.}
Although the above methods realize the editing on \shortgs, they are still limited to some simple editing operations (removal, rotation, and translation) for 3D objects.
SuGaR~\cite{SuGaR} extracts explicit meshes from the \shortgs representation by regularizing Gaussians over surfaces. 
Further, it relies on manual adjustment of Gaussian parameters based on deformed meshes to realize desired geometry editing but struggles with large-scale deformation.
SC-GS~\cite{huang2023sc} learns a set of sparse control points for 3D scene dynamics but faces challenges with intense movements and detailed surface deformation. 
GaMeS~\cite{waczynska2024games} introduces a new GS-based model that combines conventional mesh and vanilla GS. 
The explicit mesh is utilized as input and parameterizes Gaussian components using the vertices, which can modify Gaussians in real time by altering mesh components during inference.
However, it cannot handle significant deformations or changes, especially the deformation on large faces, since it cannot change the mesh topology during training.
Although the above methods can finish some simple rigid transformations and non-rigid deformation, they still face challenges in their editing effectiveness and large-scale deformation. 
As shown in Fig.~\ref{fig:edit_geo}, Gao~\etal~\cite{gao2024mesh} also adapt the mesh-based deformation to \shortgs by harnessing the priors of explicit representation (the surface properties like normals of the mesh, and the gradients generated by explicit deformation methods) and learning the face split to optimize the parameters and number of Gaussians, which provides adequate topological information to \shortgs and improves the quality for both the reconstruction and geometry editing results. 
\wtmin{GaussianFrosting~\cite{GaussianFrosting} shares a similar idea as Gao~\etal~\cite{gao2024mesh} by constructing a base mesh but further develops a forsting layer to allow Gaussians to move in a small range near the mesh surface.
}

\subsection{Appearance Editing}
On the appearance side, GaussianEditor~\cite{GaussianEditor_text} proposes to first modify 2D images with language input with diffusion model~\cite{Diffusion} in the masked region generated by the recent 2D segmentation model~\cite{SAM} and updating attributes of Gaussians again similar to the previous NeRF editing work Instruct-NeRF2NeRF~\cite{Instruct-NeRF2NeRF}. 
Another independent research work also named GaussianEditor~\cite{GaussianEditor} operates similarly but it further introduces a Hierarchical Gaussian Splatting (HGS) to allow 3D editing like object inpainting. 
\wtmin{GSEdit~\cite{GSEdit} takes a texture mesh or pre-trained \shortgs as input and utilizes the Instruct-Pix2Pix~\cite{InstructPix2Pix} and the SDS loss to updated the input mesh or \shortgs.
To alleviate the inconsistency issue, GaussCtrl~\cite{GaussCtrl} introduces the depth map as the conditional input of the ControlNet~\cite{ControlNet} to encourage geometry consistency.
Wang~\etal~\cite{wang2024viewconsistent} also aims to solve this inconsistent issue by introducing multi-view cross-attention maps.
Texture-GS~\cite{Texture-GS} proposes to disentangle the geometry and appearance of \shortgs and learns a UV mapping network for points near the underlying surface, thus enabling manipulations including texture painting and texture swapping.
3DGM~\cite{3DGM} also represents a 3D scene with a proxy mesh with fixed UV mapping where Gaussians are stored on the texture map. 
This disentangled representation also allows animation and texture editing.
Apart from local texture editing, there are works~\cite{StyleGaussian,GSinStyle,StylizedGS} focusing on stylizing \shortgs with a reference style image.
}

\begin{figure}[t]
    \centering
    \includegraphics[width=0.99\linewidth]{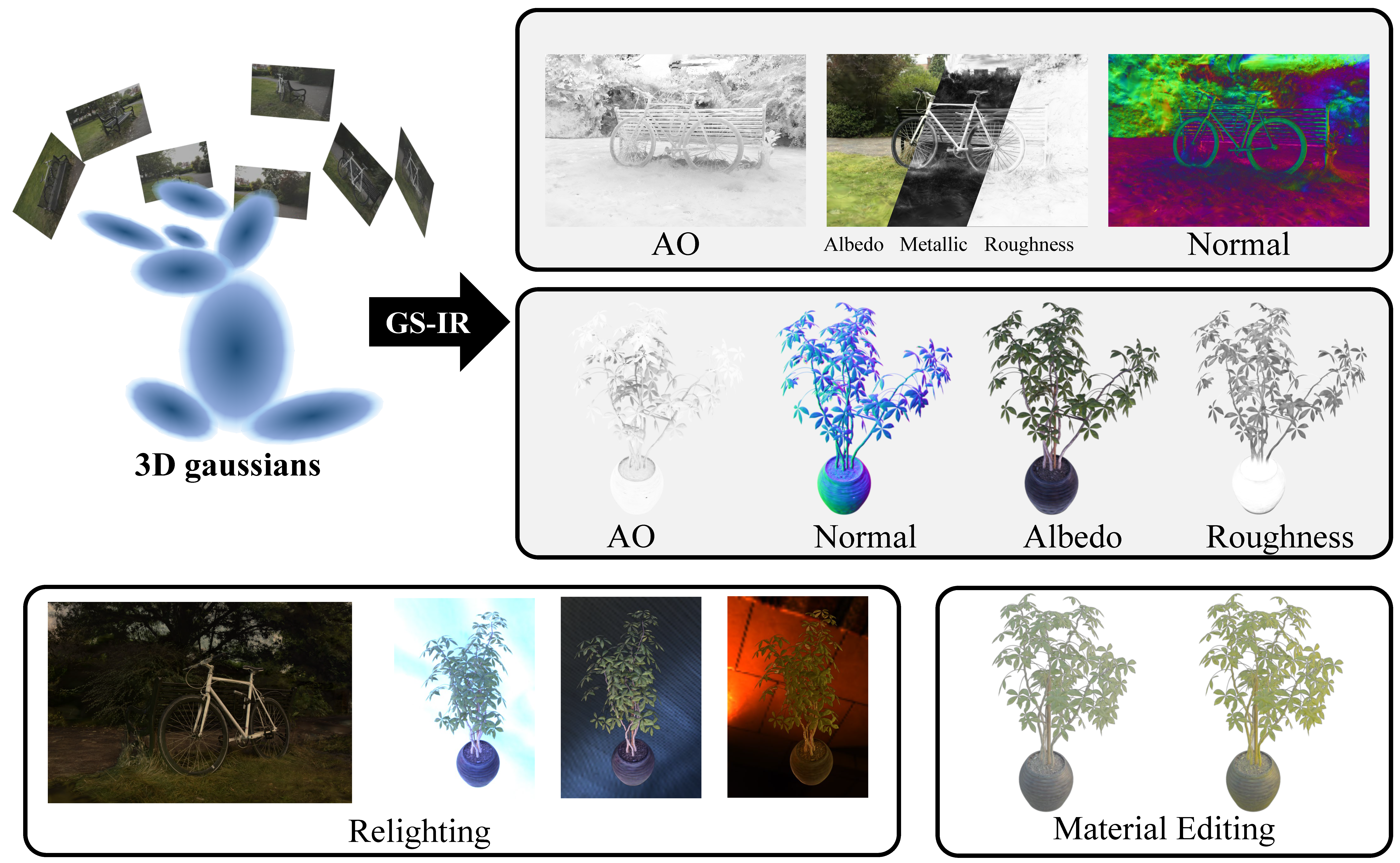}
    \caption{By decomposing material and lighting, GS-IR~\cite{GS-IR} enables appearance editing including relighting and material manipulation.}
    \label{fig:edit_appearance}
\end{figure}

To allow more tractable control over texture and lighting, researchers have started to disentangle texture and lighting to enable independent editing. 
As shown in Fig.~\ref{fig:edit_appearance}, GS-IR~\cite{GS-IR} and RelightableGaussian~\cite{RelightableGaussian} separately model texture and lighting. 
Additional materials parameters are defined on each Gaussian to represent texture and lighting is approximated by a learnable environment map. 
GIR~\cite{GIR} and GaussianShader~\cite{GaussianShader} share the same disentanglement paradigm by binding material parameters onto 3D Gaussians, but to deal with more challenging reflective scenes, they add normal orientation constraints to Gaussians similar to Ref-NeRF~\cite{Ref-NeRF}. 
After texture and lighting disentanglement, these methods can independently modify texture or lighting without influencing the other. 

\subsection{Physical Simulation}
\begin{figure*}[h]
    \centering
    \includegraphics[width=0.99\linewidth]{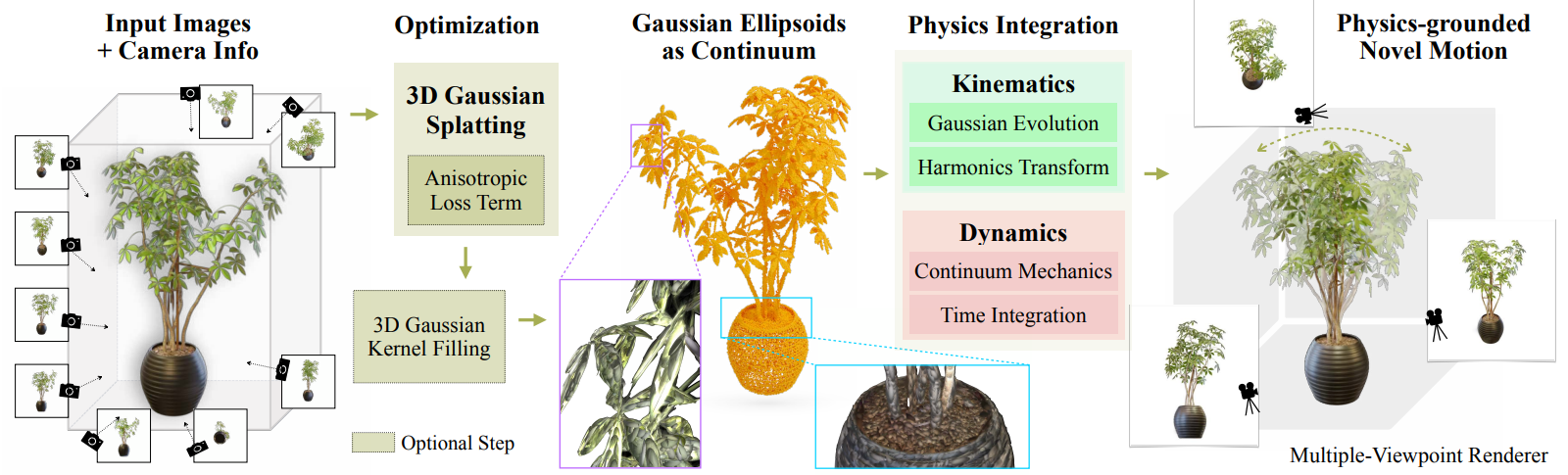}
    \caption{Pipeline of PhysGaussian~\cite{xie2023physgaussian}. 
    Treating 3D Gaussians as continuum, PhysGaussian~\cite{xie2023physgaussian} produces realistic physical simulation results.}
    \label{fig:edit_physical_simulation}
\end{figure*}

On the physical-based \shortgs editing, 
as shown in Fig.~\ref{fig:edit_physical_simulation}, PhysGaussian~\cite{xie2023physgaussian} employs discrete particle clouds from 3D GS for physically-based dynamics and photo-realistic rendering through continuum deformation~\cite{bonet1997nonlinear} of Gaussian kernels. 
Gaussian Splashing~\cite{feng2024gaussian} combines \shortgs and position-based dynamics (PBD)~\cite{macklin2016xpbd} to manage rendering, view synthesis, and solid/fluid dynamics cohesively. Similar to Gaussian shaders~\cite{GaussianShader}, the normal is applied to each Gaussian kernel to align its orientation with the surface normal and improve PBD simulation, also allowing the physically-based rendering to enhance dynamic surface reflections on fluids.
VR-GS~\cite{jiang2024vr} is a physical dynamics-aware interactive Gaussian Splatting system for VR, tackling the difficulty of editing high-fidelity virtual content in real time. 
VR-GS utilizes \shortgs to close the quality gap between generated and manually crafted 3D content. 
By utilizing physically-based dynamics, which enhance immersion and offer precise interaction and manipulation controllability. 
\wtmin{Spring-Gaus~\cite{Spring-Gaus} applies the Spring-Mass model into the modeling of dynamic \shortgs and learns the physical properties like mass and velocity from input video, which are editable for real-world simulation.
Feature Splatting~\cite{FeatureSplatting} further incorporates semantic priors from pre-trained networks and makes object-level simulation possible.}

\section{Applications of Gaussian Splatting}
\label{sec:applications}
\subsection{Segmentation and Understanding}

Open-world 3D scene understanding is an essential challenge for robotics, autonomous driving, and VR/AR environments.
With the remarkable progress in 2D scene understanding brought by SAM~\cite{SAM} and its variants, existing methods have tried to integrate semantic features, such as CLIP~\cite{CLIP}/DINO~\cite{DINO} into NeRF, to deal with 3D segmentation, understanding, and editing.

NeRF-based methods are computationally intensive because of the implicit and continuous representation.
Recent methods try to integrate 2D scene understanding methods with 3D Gaussians to produce a real-time and easy-to-editing 3D scene representation.
Most methods utilize pre-trained 2D segmentation methods like SAM~\cite{SAM} to produce semantic masks of input multi-view images~\cite{huang2023pointn,GaussianGrouping,cen2023segment,zhou2023feature,qin2023langsplat,hu2024segment,SemanticGaussians,Gaga}, or extract dense language features, CLIP~\cite{CLIP}/DINO~\cite{DINO}, of each pixel~\cite{shi2023language,zuo2024fmgs,dou2024cosseggaussians}.

LEGaussians~\cite{shi2023language} adds an uncertainty value attribute and semantic feature vector attribute for each Gaussian.
It then renders a semantic map with uncertainties from a given viewpoint, to compare with the quantized CLIP and DINO dense features of the ground truth image.
To achieve the 2D mask consistency across views, Gaussian Grouping~\cite{GaussianGrouping} employs DEVA to propagate and associate masks from different viewpoints.
It adds Identity Encoding attributes to 3D Gaussians and renders the identity feature map to compare with the extracted 2D masks.

\begin{figure}[t]
    \centering
    \includegraphics[width=0.99\linewidth]{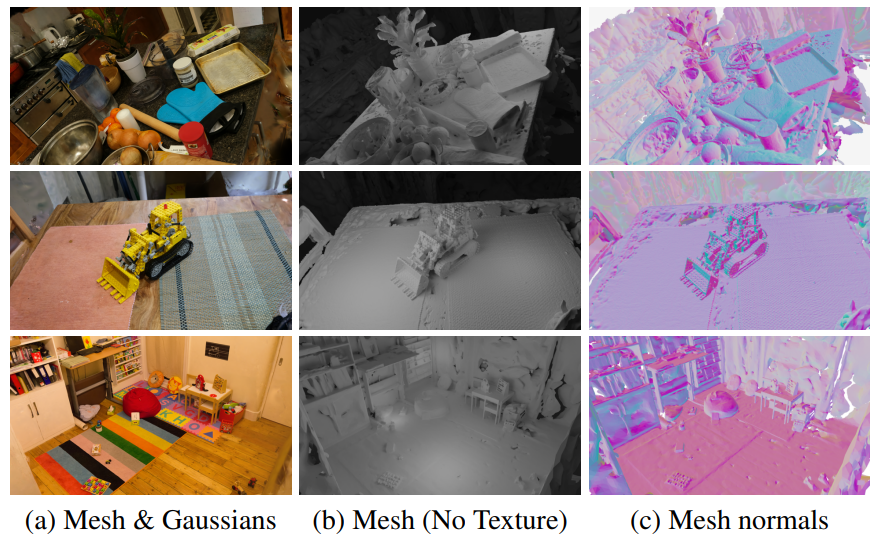}
    \caption{Geometry reconstruction results by SuGaR~\cite{SuGaR}.}
    \label{fig:geo_recon}
\end{figure}

\subsection{Geometry Reconstruction and SLAM}
Geometry reconstruction and SLAM are important subtasks in 3D reconstruction.

\paragraph{Geometry reconstruction} In the context of NeRF, a series of works~\cite{NeuS,NeUDF,LoDNeuS,Ref-NeuS} have successfully reconstructed high-quality geometry from multi-view images.
However, due to the discrete nature of \shortgs, only a few works stepped into this field. 
SuGaR~\cite{SuGaR} is the pioneering work that builds up 3D surfaces from multi-view images with the \shortgs representation.
It introduces a simple but effective self-regularization loss to constrain that the distance between the camera and the closest Gaussian should be as close as possible to the corresponding pixel's depth value in the rendered depth map, which encourages the alignment between \shortgs and the authentic 3D surface.
Another work NeuSG~\cite{NeuSG} chooses to incorporate the previous NeRF-based surface reconstruction method NeuS~\cite{NeuS} in the \shortgs representation to transfer the surface property to \shortgs.
More specifically, it encourages Gaussians' signed distance values to be zeros and the normal directions of \shortgs and the NeuS method to be as consistent as possible.
\wtmin{3DGSR~\cite{3DGSR} and GSDF~\cite{GSDF} also encourage the consistency between SDF and \shortgs to enhance the geometry reconstruction quality.
DN-Splatter~\cite{DN-Splatter} utilizes depth and normal priors captured from common devices or predicted from general-purpose networks to enhance the reconstruction quality of \shortgs.
Wolf~\etal~\cite{wolf2024surface} first train a \shortgs to render stereo-calibrated novel views and apply stereo depth estimation on the rendered views. 
The estimated dense depth maps are fused together by the Truncated Signed Distance Function(TSDF) to form a triangular mesh.
2D-GS~\cite{2D-GS} replaces 3D Gaussians with 2D Gaussians for more accurate ray-splat intersection and employs a low-pass filter to avoid degenerated line projection.}
\wtmin{Though attempts have been made in the field of geometry reconstruction, due to the discrete nature of \shortgs, current methods achieve comparable or worse results compared to implicit representation-based methods with continuous field assumption where the surface can be easily determined.}

\paragraph{SLAM}There are also \shortgs methods targeting simultaneously localizing the cameras and reconstructing the 3D scenes. 
GS-SLAM~\cite{GS-SLAM} proposes an adaptive 3D Gaussian expanding strategy to add new 3D Gaussians into the training stage and delete unreliable ones with captured depths and rendered opacity values.
To avoid duplicate densification, SplaTAM~\cite{SplaTAM} uses view-independent colors for Gaussians and creates a densification mask to determine whether a pixel in a new frame needs densification by considering current Gaussians and the captured depth of the new frame.
For stabilizing the localization and mapping, GaussianSplattingSLAM~\cite{GaussianSplattingSLAM} and Gaussian-SLAM~\cite{Gaussian-SLAM} put an extra scale regularization loss on the scale of Gaussians to encourage isotropic Gaussians.
For easier initialization, LIV-GaussMap~\cite{LIV-GaussMap} initializes Gaussians with LiDAR point cloud and builds up an optimizable size-adaptive voxel grid for the global map.
SGS-SLAM~\cite{SGS-SLAM}, NEDS-SLAM~\cite{NEDS-SLAM}, and SemGauss-SLAM~\cite{SemGauss-SLAM} further consider Gaussian's semantic information in the simultaneous localization and mapping process by distilling 2D semantic information which can be obtained using 2D segmentation methods or provided by the dataset.
Deng~\etal~\cite{deng2024compact} avoid redundant Gaussian splitting based on sliding window mask and use vector quantization to further encourage compact \shortgs.
CG-SLAM~\cite{CG-SLAM} introduces an uncertainty map into the training process based on the rendered depth and greatly improves the reconstruction quality. 
Based on the reconstructed map by SLAM-based methods, tasks in robotics like relocalization~\cite{3DGS-ReLoc}, navigation~\cite{Splat-Nav,GaussNav,liu2024uncertainty}, 6D pose estimation~\cite{GS-Pose}, multi-sensor calibration~\cite{MM3DGSSLAM,3DGS-Calib} and manipulation~\cite{ManiGaussian,GaussianGrasper} can be performed efficiently. 
We report quantitative results by different SLAM methods on the reconstruction task in Table.~\ref{tab:SLAM}. 
\wtmin{The explicit geometry representation provided by \shortgs enables flexible reprojection to alleviate the misalignment between different viewpoints, thus leading to better reconstruction compared to NeRF-based methods.
The real-time rendering feature of \shortgs also makes neural-based SLAM methods more applicable as the training in previous NeRF-based methods require more hardware and time.}

\begin{table}[t!]
    \caption{
        Quantitative comparison of novel view synthesis results by different SLAM methods on the Replica~\cite{Replica} dataset using PSNR, SSIM and LPIPS metrics.
    }
        \centering
                \setlength\tabcolsep{0pt}
                \begin{tabular*}{\columnwidth}{@{\extracolsep{\fill}} cccc}
\hline
                    Methods                                            & PSNR $\uparrow$ & SSIM $\uparrow$ & LPIPS $\downarrow$\\ \hline
                    NICE-SLAM~\cite{NICE-SLAM}                         & 24.42           & 0.81            & 0.23              \\
                    Vox-Fusion~\cite{Vox-Fusion}                       & 24.41           & 0.80            & 0.24              \\
                    Co-SLAM~\cite{Co-SLAM}                             & 30.24           & 0.94            & 0.25              \\ \hline
                    GS-SLAM~\cite{GS-SLAM}                             & 31.56           & \sbc{0.97}      & \tbc{0.094}       \\
                    SplaTAM~\cite{SplaTAM}                             & 34.11           & \sbc{0.97}      & 0.10              \\
                    GaussianSplattingSLAM~\cite{GaussianSplattingSLAM} & \sbc{37.50}     & \tbc{0.96}      & \bc{0.07}         \\
                    Gaussian-SLAM~\cite{Gaussian-SLAM}                 & \bc{38.90}      & \bc{0.99}       & \bc{0.07}         \\
                    SGS-SLAM~\cite{SGS-SLAM}                           & \tbc{34.15}     & \sbc{0.97}      & 0.096             \\ \hline
                \end{tabular*}
    \label{tab:SLAM}
\end{table}

\def\eg{\textit{e.g.\xspace}}
\subsection{Digital Human}
Learning virtual humans with implicit representation has been explored in various ways, especially for the NeRF and SDF representations, which exhibit high-quality results from multi-view images but suffer from heavy computational costs.
Thanks to the high efficiency of \shortgs, research works have flourished and pushed \shortgs into digital human creation.

\paragraph{Human body} In full-body modeling, works aim to reconstruct dynamic humans from multi-view videos.
D3GA~\cite{zielonka2023drivable} first creates animatable human avatars using drivable 3D Gaussians and tetrahedral cages, which achieves promising geometry and appearance modeling.
To capture more dynamic details, SplatArmor~\cite{jena2023splatarmor} leverages two different MLPs to predict large motions built upon the SMPL and canonical space and allows the pose-dependent effects by the proposed SE(3) fields, enabling more detailed results.
HuGS~\cite{moreau2023human} creates a coarse-to-fine deformation module using linear blend skinning and local learning-based refinement for constructing and animating virtual human avatars based on \shortgs.
It achieves state-of-the-art human neural rendering performance at 20 FPS.
Similarly, HUGS~\cite{kocabas2023hugs} utilizes the tri-plane representation~\cite{chan2022efficient} to factorize the canonical space, which can reconstruct the person and scene from monocular video (50~100 frames) within 30 minutes.
Since \shortgs learns a huge number of Gaussians ellipsoids, HiFi4G~\cite{jiang2023hifi4g} combines \shortgs with the non-rigid tracking offered by its dual-graph mechanism for high-fidelity rendering, which successfully preserves spatial-temporal consistency in a more compact manner.
To achieve higher rendering speeds with high resolution on consumer-level devices,
GPS-Gaussian~\cite{zheng2023gps} introduces Gaussian parameter maps on the sparse source view to regress the Gaussian parameters jointly with a depth estimation module without any fine-tuning or optimization.
Other than that, GART~\cite{lei2023gart} extends the human to more articulated models (\eg, animals) based on the \shortgs representation.

To make full use of the information from multi-view images,
Animatable Gaussians~\cite{li2023animatable} incorporates \shortgs and 2D CNNs for more accurate human appearances and realistic garment dynamics using a template-guided parameterization and pose projection mechanism.
Gaussian Shell Maps~\cite{abdal2023gaussian} (GSMs) combines CNN-based generators with \shortgs to recreate virtual humans with sophisticated details such as clothing and accessories.
ASH~\cite{pang2023ash} projects the 3D Gaussian learning into a 2D texture space using mesh UV parameterization to capture the appearance, enabling real-time and high-quality animated humans.
Furthermore, for reconstructing rich details on humans, such as the cloth, 3DGS-Avatar~\cite{qian20233dgs} introduces a shallow MLP instead of SH to model the color of 3D Gaussians and regularizes deformation with geometry priors, providing the photorealistic rendering with pose-dependent cloth deformation and generalizes to the novel poses effectively.

\begin{figure}[!t]
    \centering
    \includegraphics[width=0.99\linewidth]{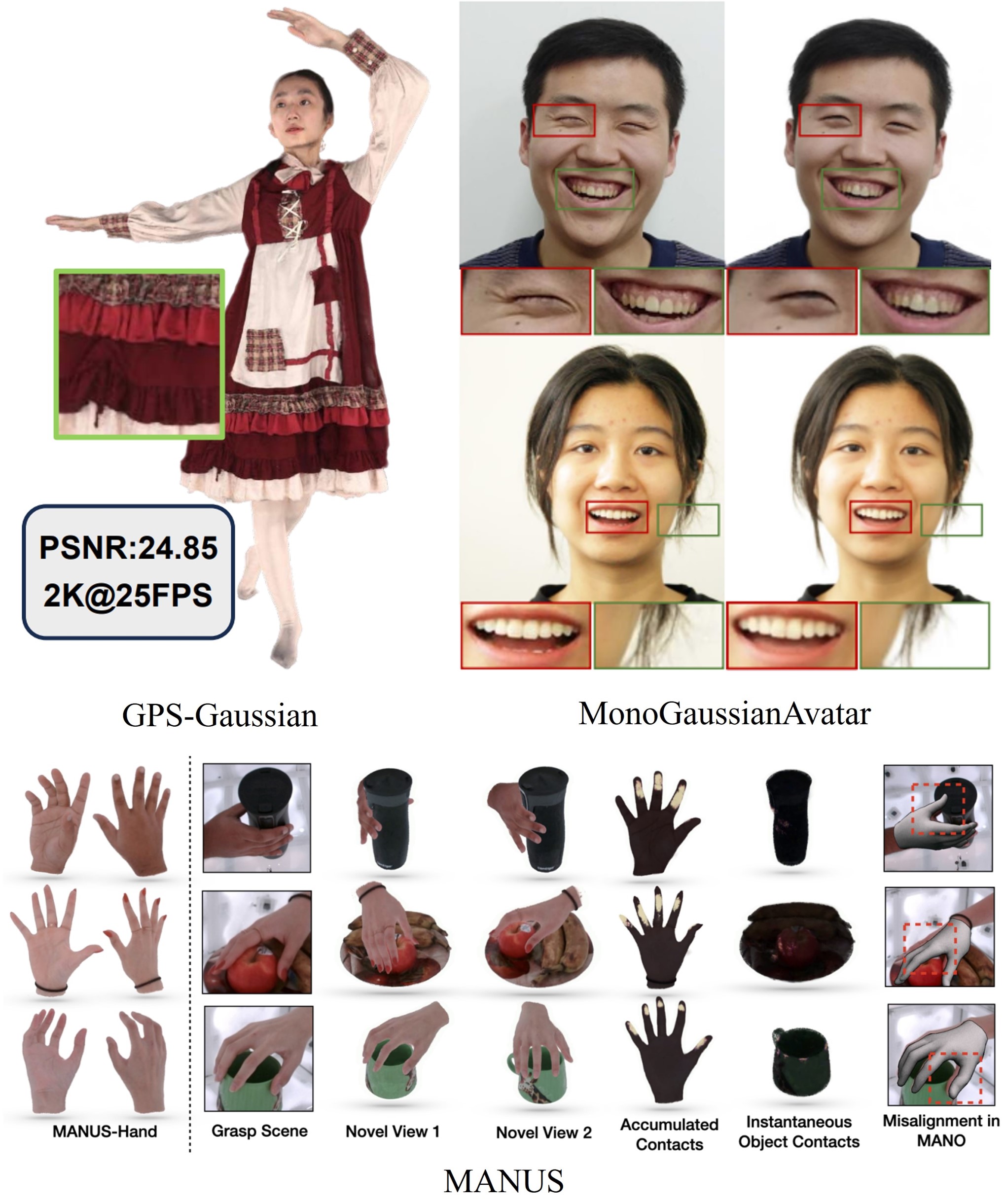}
    \caption{The results of GPS-Gaussian~\cite{zheng2023gps}, MonoGaussianAvatar~\cite{chen2023monogaussianavatar}, and MANUS~\cite{pokhariya2023manus}. They explore the 3DGS-based approaches on the whole body, head, and hand modeling, respectively.}
    \label{fig:digitalhuman}
\end{figure}

For dynamic digital human modeling based on monocular video,
GaussianBody~\cite{li2024gaussianbody} further leverages the physical-based priors to regularize the Guassians in the canonical space to avoid artifacts in the dynamic cloth from monocular video.
GauHuman~\cite{hu2023gauhuman} re-designs the prune/split/clone of the original \shortgs to achieve efficient optimization and incorporates pose refinement and weight fields modules for fine details learning.
It achieves minute-level training and real-time rendering (166 FPS).
GaussianAvatar~\cite{hu2023gaussianavatar} incorporates the optimizable tensor with a dynamic appearance network to capture the dynamics better, allowing the dynamic avatar reconstruction and realistic novel animation in real time.
Human101~\cite{li2023human101} further pushes the speed of high-fidelity dynamic human creation to 100 seconds using a fixed-perspective camera.
\wtmin{Simliar to~\cite{gao2024mesh}, SplattingAvatar~\cite{SplattingAvatar} and GoMAvatar~\cite{GoMAvatar} embed Gaussians onto a canonical human body mesh. 
The positions of a Gaussian is determined by the barycentric point and the displacement along the normal direction.
To resolve the unbalanced aggregation of Gaussians caused by densification and splitting operation, GVA~\cite{GVA} proposes a surface-guided Gaussians re-initialization strategy to make the trained Gaussians better fit the input monocular video.
HAHA~\cite{HAHA} also attaches Gaussians onto the surface of a mesh but combines rendered results from a textured human body mesh and Gaussians together to reduce the number of Gaussians.
}

\paragraph{Head} For human head modeling with \shortgs,
MonoGaussianAvatar~\cite{chen2023monogaussianavatar} first applies \shortgs to dynamic head reconstruction using the canonical space modeling and deformation prediction.
Further,
PSAvatar~\cite{zhao2024psavatar} introduces the explicit Flame face model~\cite{FLAME:SiggraphAsia2017} to initialize Gaussians, which can capture the high-fidelity facial geometry and even complicated volumetric objects (\eg glasses).
The tri-plane representation and the motion fields are used in GaussianHead~\cite{wang2023gaussianhead} to simulate geometrically changing heads in continuous movements and render rich textures, including the skin and hair.
For easier head expression controllability,
GaussianAvatars~\cite{qian2023gaussianavatars} introduce the geometric priors (Flame parametric face model~\cite{FLAME:SiggraphAsia2017}) into \shortgs, which binds the Gaussians onto the explicit mesh and optimize the parameters of Gaussian ellipsoids.
Rig3DGS~\cite{rivero2024rig3dgs} employs a learnable deformation to provide stability and generalization to novel expressions, head poses, and viewing directions to achieve controllable portraits on portable devices.
In another way, HeadGas~\cite{dhamo2023headgas} attributes the \shortgs with a base of latent features that are weighted by the expression vector from 3DMMs~\cite{blanz2023morphable}, which achieves real-time animatable head reconstruction.
FlashAvatar~\cite{xiang2023flashavatar} further embeds a uniform 3D Gaussian field in a parametric face model and learns additional spatial offsets to capture facial details, successfully pushing the rendering speed to 300 FPS.
To synthesize the high-resolution results, Gaussian Head Avatar~\cite{xu2023gaussian} adopts the super-resolution network to achieve high-fidelity head avatar learning.
\wtmin{To synthesize high-quality avatar from few-view input, SplatFace~\cite{SplatFace} propose to first initialize Gaussians on a template mesh and jointly optimizing Gaussians and the mesh with a splat-to-mesh distance loss. 
GauMesh~\cite{GauMesh} propose a hybrid representation containing both tracked textured meshes and canonical 3D Gaussians together with a learnable deformation field to represent dynamic human head.}
Apart from these, some works extend the \shortgs into text-based head generation~\cite{zhou2024headstudio}, DeepFake~\cite{stanishevskii2024implicitdeepfake}, and relighting~\cite{saito2023relightable}.

\paragraph{Hair and hands} Other parts of humans have also been explored, such as hair and hands.
3D-PSHR~\cite{jiang20233d} combines hand geometry priors (MANO) with \shortgs, which first realizes the real-time hand reconstruction.
MANUS~\cite{pokhariya2023manus} further explores the interaction between the hands and object using \shortgs.
In addition, GaussianHair~\cite{luo2024gaussianhair} first combines the Marschner Hair Model~\cite{marschner2003light} with UE4's real-time hair rendering to create the Gaussian Hair Scattering Model.
It captures complex hair geometry and appearance for fast rasterization and volumetric rendering, enabling applications including editing and relighting.

\subsection{3D/4D Generation}
Cross-modal image generation has achieved stunning results with the diffusion model~\cite{Diffusion}.
However, due to the lack of 3D data, it is difficult to directly train a large-scale 3D generation model.
The pioneering work DreamFusion~\cite{poole2022dreamfusion} exploits the pre-trained 2D diffusion model and proposes the score distillation sampling (SDS) loss, which distills the 2D generative priors into 3D without requiring 3D data for training, achieving text-to-3D generation.
However, the NeRF representation brings heavy rendering overhead.
The optimization time for each case takes several hours and the rendering resolution is low, which leads to poor-quality results.
Although some improved methods extract mesh representation from trained NeRF for fine-tuning to improve the quality~\cite{lin2023magic3d}, this way will further increase optimization time.
\wtmin{\shortgs representation can render high-resolution images with high FPS and small memory, so it replaces NeRF as the 3D representation in some recent 3D/4D generation methods.}

\begin{figure}[!t]
    \centering
    \includegraphics[width=0.99\linewidth]{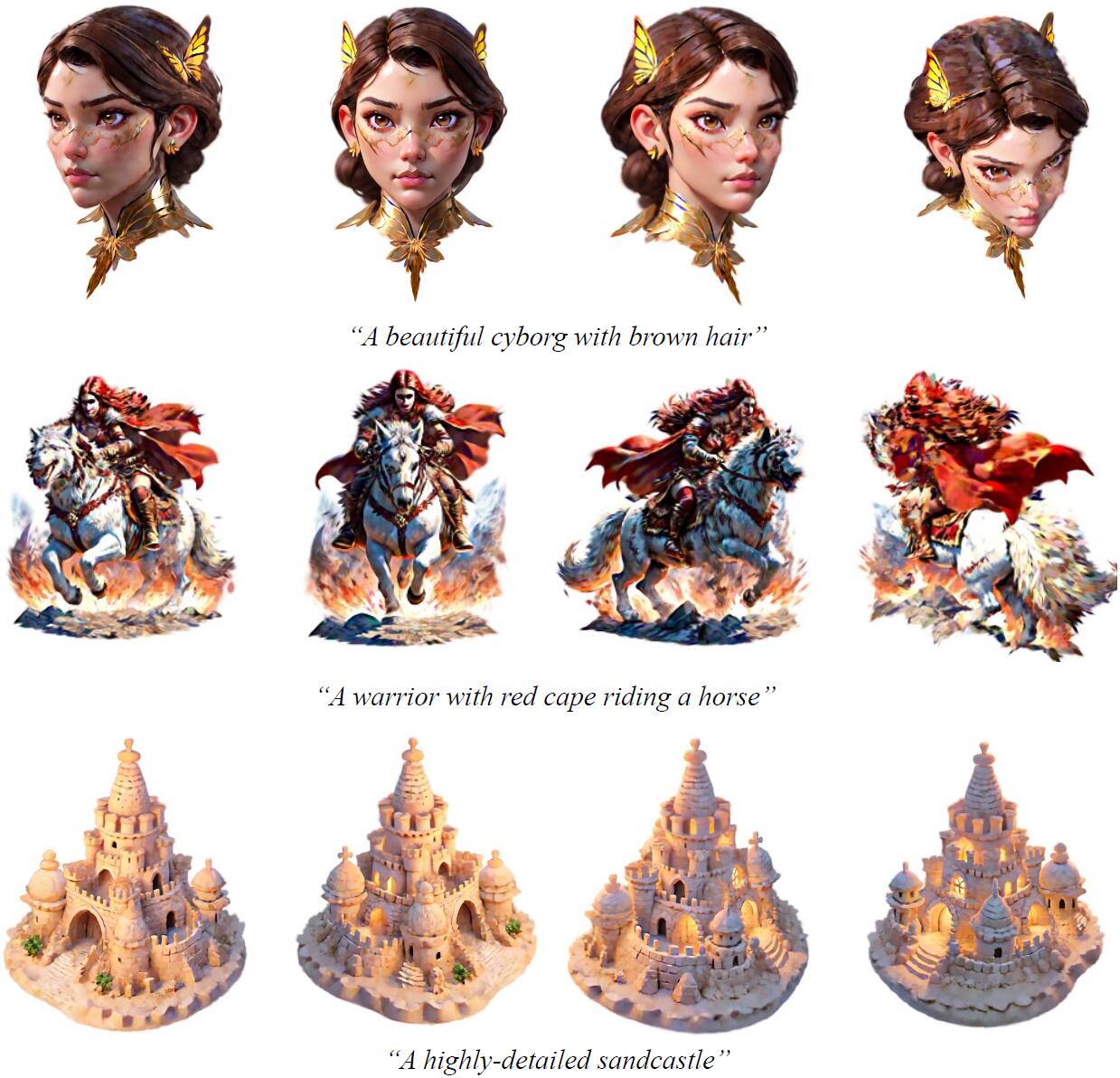}
    \caption{The text-to-3D generation results of Luciddreamer~\cite{liang2023luciddreamer}. It distills generative prior from pre-trained 2D diffusion models with the proposed Interval Score Matching (ISM) objective to achieve 3D generation from the text prompt.}
    \label{fig:3dgen}
\end{figure}

\paragraph{3D generation} DreamGaussian~\cite{tang2023dreamgaussian} replaces the MipNeRF~\cite{barron2021mip} representation in the DreamFusion~\cite{poole2022dreamfusion} framework with \shortgs that uses SDS loss to optimize 3D Gaussians.
The splitting process of \shortgs is very suitable for the optimization progress under the generative settings, so the efficiency advantages of \shortgs can be brought to text-to-3D generation based on the SDS loss.
To improve the final quality, this work follows the idea of Magic3D~\cite{lin2023magic3d} which extracts the mesh from the generated \shortgs and refines the texture details by optimizing UV textures through a pixel-wise Mean Squared Error (MSE) loss.
In addition to 2D SDS, GSGEN~\cite{chen2023text} introduces a 3D SDS loss based on Point-E~\cite{nichol2022point}, a text-to-point-cloud diffusion model, to mitigate the multi-face or Janus problem.
It adopts Point-E to initialize the point cloud as the initial geometry for optimization and also refines the appearance with only the 2D image prior.
GaussianDreamer~\cite{yi2023gaussiandreamer} also combines the priors of 2D and 3D diffusion models.
It utilizes Shap-E~\cite{jun2023shap} to generate the initial point cloud and optimizes \shortgs using 2D SDS.
However, the generated initial point cloud is relatively sparse, so noisy point growth and color perturbation are further proposed to densify it.
However, even if the 3D SDS loss is introduced, the Janus problem may still exist during optimization as the view is sampled one by one.
Some methods~\cite{shi2023mvdream, Wang2023ImageDreamIM} fine-tune the 2D diffusion model~\cite{Diffusion} to generate multi-view images at once, thereby achieving multi-view supervision during SDS optimization. 
Or, the multi-view SDS proposed by BoostDream~\cite{yu2024boostdream} directly creates a large $2 \times 2$ images by stitching the rendered images from 4 sampled views and calculates the gradients under the condition of the multi-view normal map.
This is a plug-and-play method that can first convert a 3D asset into differentiable representations including NeRF, \shortgs, and DMTet~\cite{shen2021deep} through rendering supervision, and then optimize them to improve the quality of the 3D asset.
Some methods have made improvements to SDS loss.
LucidDreamer~\cite{liang2023luciddreamer} proposes Interval Score Matching (ISM), which replaces DDPM in SDS with DDIM inversion and introduces supervisions from interval steps of the diffusion process to avoid high error in one-step reconstruction. Some generation results are shown in Fig.~\ref{fig:3dgen}.
GaussianDiffusion~\cite{li2023gaussiandiffusion} proposes to incorporate structured noises from multiple viewpoints to alleviate the Janus problem and variational \shortgs for better generation results by mitigating floaters.
Yang \etal~\cite{yang2023learn} point out that the differences between the diffusion prior and the training process of the diffusion model will impair the quality of 3D generation, so they propose iterative optimization of the 3D model and the diffusion prior.
Specifically, two additional learnable parameters are introduced in the classifier-free guidance formula, one is a learnable unconditional embedding, and the other is additional parameters added to the network, such as LoRA~\cite{hu2021lora} parameters.
These methods are not limited to \shortgs, and other originally NeRF-based methods including VSD~\cite{wang2023prolificdreamer} and CSD~\cite{yu2023text} aiming at improving the SDS loss can be used for the \shortgs generation.
\wtmin{GaussianCube~\cite{GaussianCube} instead trains a 3D diffusion model based on a GaussianCube representation that is converted from a constant number of Gaussians with  voxelization via optimal transport.
GVGEN~\cite{GVGEN} also works in 3D space but is based on a 3D Gaussian volume representation.}

As a special category, the human body can introduce the model prior, such as SMPL~\cite{loper2023smpl}, to assist in generation. GSMs~\cite{abdal2023gaussian} builds multi-layer shells from the SMPL template and binds 3D Gaussians on the shells.
By utilizing the differentiable rendering of \shortgs and the generative adversarial network of StyleGAN2~\cite{Karras2019stylegan2}, animatable 3D humans are efficiently generated.
GAvatar~\cite{yuan2023gavatar} adopts a primitive-based representation~\cite{lombardi2021mixture} attached to the SMPL-X~\cite{pavlakos2019expressive} and attaches 3D Gaussians to the local coordinate system of each primitive.
The attribute values of 3D Gaussians are predicted by an implicit network and the opacity is converted to the signed distance field through a NeuS~\cite{NeuS}-like method, providing geometry constraints and extracting detailed textured meshes.
The generation is text-based and mainly guided by the SDS loss.
HumanGaussian~\cite{liu2023humangaussian} initializes 3D Gaussians by randomly sampling points on the surface of the SMPL-X~\cite{pavlakos2019expressive} template.
It extends Stable Diffusion~\cite{Diffusion} to generate RGB and depth simultaneously and constructs a dual-branch SDS as the optimization guidance.
It also combines the classifier score provided by the null text prompt and the negative score provided by the negative prompt to construct the negative prompt guidance to address the over-saturation issue.

\begin{figure*}[!t]
    \centering
    \includegraphics[width=0.99\linewidth]{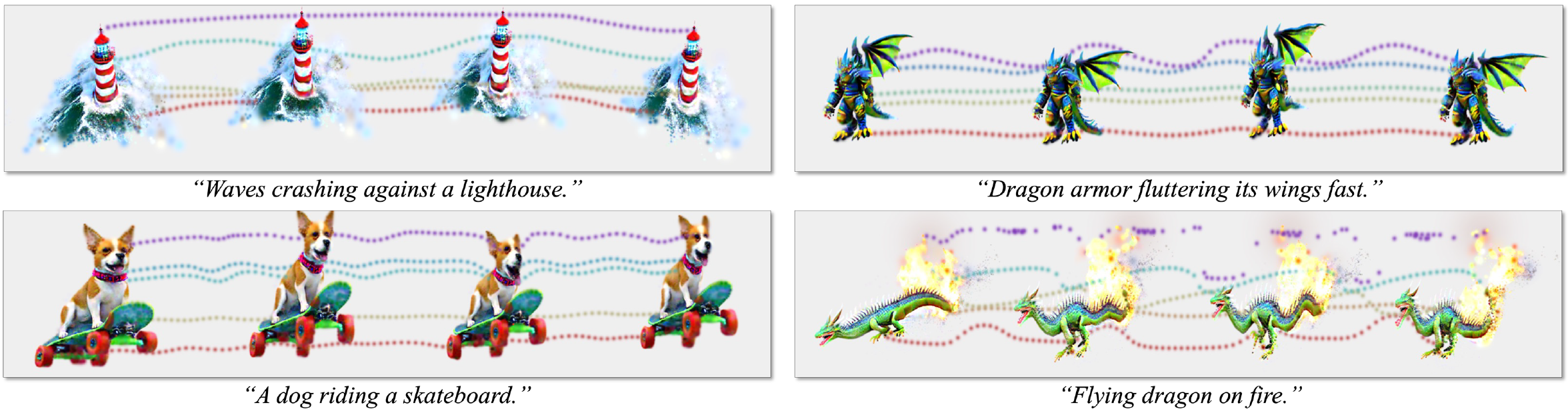}
    \caption{The text-to-4D generation results of AYG~\cite{Ling2023AlignYG}. Different dynamic 4D sequences are shown with dotted lines representing the dynamics of the deformation field.}
    \label{fig:4dgen}
\end{figure*}

The above methods focus on the generation of an individual object, while scene generation requires consideration of interactions and relationships between different objects.
CG3D~\cite{vilesov2023cg3d} inputs a text prompt manually deconstructed by the user into a scene graph, and the textual scene graph is interpreted as a probabilistic graphical model where the directed edge has a tail of the object node and a head of the interaction node.
Then scene generation becomes an ancestor sampling by first generating the objects and then their interactions.
The optimization is divided into two stages, where gravity and normal contact forces are introduced in the second stage.
LucidDreamer~\cite{chung2023luciddreamer} and Text2Immersion~\cite{ouyang2023text2immersion} are both based on a reference image (user-specified or text-generated) and extend outward to achieve 3D scene generation.
The former utilizes Stable Diffusion (SD)~\cite{Diffusion} for image inpainting to generate unseen regions on the sampled views and incorporates monocular depth estimation and alignment to establish a 3D point cloud from these views.
Finally, the point cloud is used as the initial value, and a \shortgs is trained using the projected images as the ground truth to achieve 3D scene generation.
The latter method has a similar idea, while there is a process to remove outliers in the point cloud and the \shortgs optimization has two stages: training coarse \shortgs and refinement. 
\wtmin{To explore 3D scene generation, GALA3D~\cite{GALA3D} leverages both the object-level text-to-3D model MVDream~\cite{shi2023mvdream} to generate realistic objects and scene-level diffusion model to compose them.
DreamScene~\cite{DreamScene} proposes multi-timestep sampling for multi-stage scene-level generation to synthesize surrounding environment, ground, and objects to avoid complicate object composition.
Instead of separately modeling each object, RealmDreamer~\cite{RealmDreamer} utilizes the diffusion priors from inpainting and depth estimation models to generate different viewpoints of a scene iteratively. 
DreamScene360~\cite{DreamScene360} instead generates a 360-degree panoramic image and converts it into a 3D scene with depth estimation.}

Text-to-3D generation methods can be applied to image-to-3D generation, or monocular 3D reconstruction, with some simple modifications.
For example, the pre-trained diffusion model used in SDS loss can be replaced with Zero-1-to-3 XL~\cite{liu2023zero} for image condition~\cite{tang2023dreamgaussian}.
We can also add losses between the input image and the corresponding rendered image under the input view to make the generation more consistent with the input image.
Based on the image-to-3D generation of DreamGaussian~\cite{tang2023dreamgaussian}, Repaint123~\cite{zhang2023repaint123} proposes a progressive controllable repainting mechanism to refine the generated mesh texture. During the process of repainting the occlusions, it incorporates textural information from the reference image through attention feature injection~\cite{cao2023masactrl} and proposes a visibility-aware repainting process to refine the overlap regions with different strengths.
Finally, the refined images will be used as ground truths to directly optimize the texture through MSE loss, achieving fast optimization.
Other methods explore utilizing existing 3D datasets~\cite{deitke2023objaverse,deitke2023objaversexl} and constructing large models to directly generate \shortgs representation from a single image.
TriplaneGaussian~\cite{zou2023triplane} proposes a hybrid representation of tri-plane and \shortgs.
It generates a point cloud and a tri-plane encoding \shortgs's attributes information from the input image features through a transformer-based point cloud decoder and a tri-plane decoder, respectively.
The generated point cloud is densified through an upsampling method and then projected onto the tri-plane to query features.
The queried features are augmented by the projected image features and are translated into 3D Gaussian attributes using an MLP, thereby achieving the generation of \shortgs from a single image.
LGM~\cite{tang2024lgm} first exploits the off-the-shelf models to generate multi-view images from text~\cite{shi2023mvdream} or a single image~\cite{Wang2023ImageDreamIM}.
Then it trains a U-Net-based network to generate \shortgs from the multi-view images.
The U-Net is asymmetric, which allows for the input of high-resolution images while limiting the number of output Gaussians.
AGG~\cite{xu2024agg} also introduces a hybrid generator to obtain the point cloud and tri-plane features.
But it first generates a coarse \shortgs and then upsamples it through a U-Net-based super-resolution module to improve the fidelity of generated results.
\wtmin{Instead of predicting point cloud from image, BrightDreamer~\cite{BrightDreamer} predicts the deviation of a set of fixed anchor points to determine the centers of Gaussians.
GRM~\cite{GRM} utilizes existing multi-view generation models to train a pixel-aligned Gaussian representation for faithful 3D generation via a single feed-forward pass.
IM-3D~\cite{IM-3D} finetunes the image-to-video model based on Emu~\cite{Emu} to generate a turn-table like video rotating around an object, which are taken as the input for \shortgs reconstruction. 
Gamba~\cite{Gamba} proposes to predict Gaussian attributes with the recent Mamba~\cite{Mamba} network to better capture the relationship between Gaussians.
MVControl~\cite{MVControl} extends ControlNet~\cite{ControlNet} to 3D generation task and allows extra condition inputs like edge, depth, normal, and scribble fed into existing multi-view generation models.
Hyper-3DG~\cite{Hyper-3DG} proposes a geometry and texture refinement module to improve generation quality based on hypergraph learning where each node is a patch image of a coarse \shortgs.
For the same purpose, DreamPolisher~\cite{DreamPolisher} utilizes a ControlNet-based network for texture refinement and ensures consistency between different viewpoints with a view-consistent geometric guidance.
FDGaussian~\cite{FDGaussian} injects tri-plane features into the diffusion model for better geometry generation.}

\paragraph{4D generation} Based on the current progress of 3D generation, preliminary exploration has also been performed on 4D generation with \shortgs representation.
AYG~\cite{Ling2023AlignYG} endows \shortgs with dynamics with a deformation network for the text-to-4D generation.
It is divided into two stages, static \shortgs generation with the SDS losses based on Stable Diffusion~\cite{Diffusion} and MVDream~\cite{shi2023mvdream}, and then dynamic generation with a video SDS loss based on a text-to-video diffusion model~\cite{blattmann2023align}.
In the dynamic generation stage, only the deformation field network is optimized, and some frames are randomly selected to add image-based SDS to ensure generation quality. The generation results are shown in Fig.~\ref{fig:4dgen}.
DreamGaussian4D~\cite{ren2023dreamgaussian4d} achieves 4D generation given a reference image.
A static \shortgs is first generated using an improved version of DreamGaussian~\cite{tang2023dreamgaussian}.
The off-the-shelf Stable Diffusion Video is utilized to generate a video from the given image.
Then the dynamic generation is also realized by the optimization of a deformation network added to the static \shortgs, and the generated video is used as supervision, along with a 3D SDS loss based on Zero-1-to-3 XL~\cite{liu2023zero} from sampled views.
Finally, this method also extracts a mesh sequence and optimizes the texture with an image-to-video diffusion model.
Last, for video-to-4D generation, 4DGen~\cite{yin20234dgen} and Efficient4D~\cite{pan2024fast} both propose utilizing SyncDreamer~\cite{liu2023syncdreamer} to generate multi-view images from the input frames as pseudo ground truth to train a dynamic \shortgs.
The former introduces HexPlane~\cite{cao2023hexplane} as the dynamic representation and constructs point clouds using generated multi-view images as 3D deformation pseudo ground truth.
The latter directly converts 3D Gaussian into 4D Gaussian and enhances the temporal continuity of SyncDreamer~\cite{liu2023syncdreamer} by fusing spatial volumes at adjacent timestamps, achieving time synchronization to generate better cross-time multi-view images for supervision.
\wtmin{SC4D~\cite{SC4D} migrates the idea of sparse control point in SC-GS~\cite{huang2023sc} to model deformation and appearance more efficiently.
STAG4D~\cite{STAG4D} proposes a temporally consistent multi-view diffusion model to generate multi-view videos for 4D reconstruction from monocular video or 4D generation. 
To overcome previous 4D generation models' object-centric limitation, Comp4D~\cite{Comp4D} first generates individual 4D objects and later composes them under trajectory constraints.
}
\wtmin{To enable more realistic 3D/4D generation, most methods utilize the priors from diffusion models, which operate on 2D images and require a rendered viewpoint of the generated object. 
The fast rasterization-based rendering in \shortgs makes the priors more efficiently applied compared to NeRF-based methods with slow rendering speed.}

\section{Conclusions and Discussions}
\label{sec:conclusion}
\begin{figure*}[!t]
    \centering
    \includegraphics[width=0.99\linewidth]{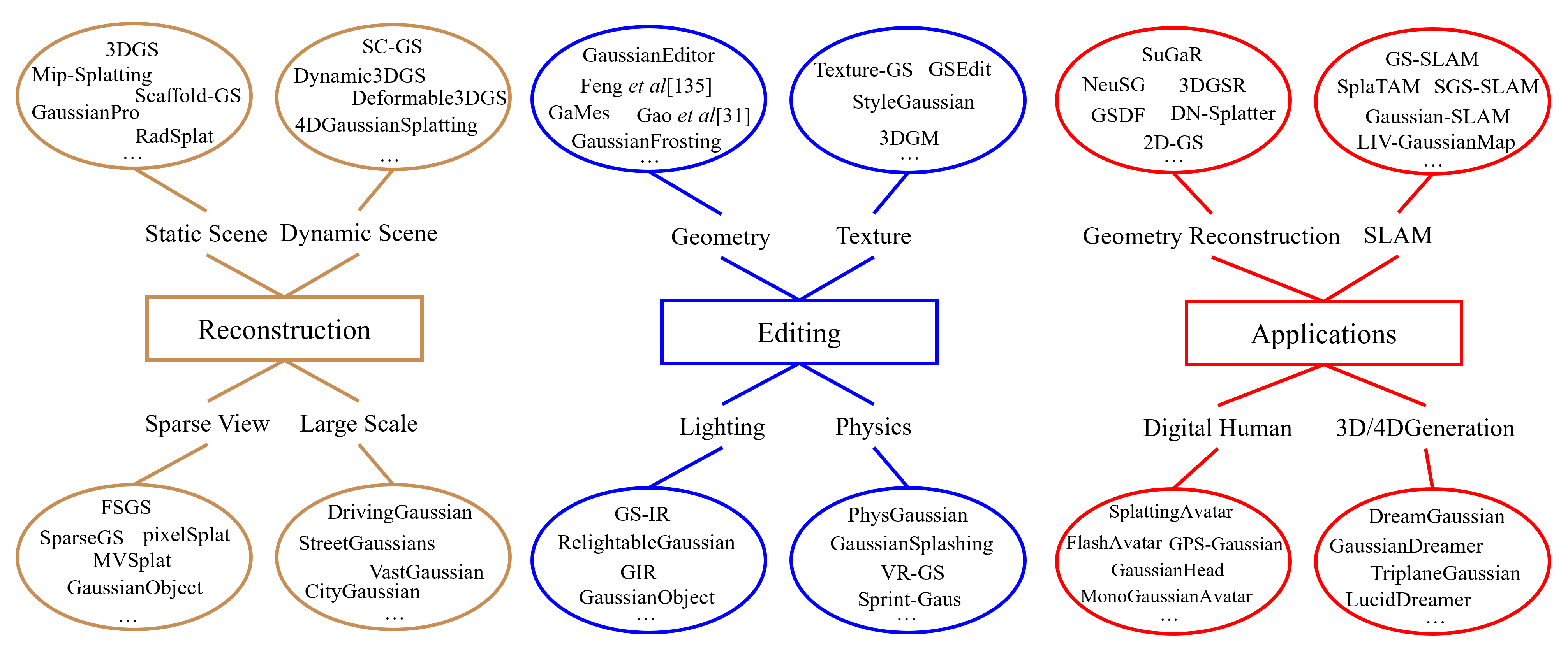}
    \caption{Representative works based on the \shortgs representation across different tasks.}
    \label{fig:tree}
\end{figure*}

This survey presents an overview of the recent \longgs (\shortgs) technique, which not only illustrates how it originates from traditional point-based rendering methods but also how its fast rendering and explicit geometry facilitate a series of representative works targeting different tasks like 3D reconstruction and 3D editing as shown in the tree diagram in Fig.~\ref{fig:tree}. 

\wtmin{
Although \shortgs has greatly improved the efficiency and results' quality on a few tasks, it is not a perfect 3D representation that can satisfy the need of all tasks. 
Here we first discuss the advantages and disadvantages of commonly used 3D representations including mesh, SDF, NeRF, and \shortgs. 
}

\wtmin{
\textbf{Mesh.}
Mesh is made of a set of vertices, edges, and faces, which can express detailed geometry at relatively lower storage cost. 
As the most widely used 3D representation in the industrial, it can create high-quality visual effects with the help of physically-based materials. in real-time 
However, most meshes are made by artists or creators, which takes a lot of time. 
Even if some works~\cite{TM-NET,SDM-NET,PolyGen,MeshGPT} attempt to generate meshes automatically with neural networks, their generation capability is still constrained by existing datasets with limited quantity. 
}

\wtmin{
\textbf{SDF and NeRF.}
Both SDF and NeRF are based on the implicit neural fields, which can be learned from a set of multi-view images automatically. 
The explicit geometry or mesh can also be extracted from SDF or NeRF with the marching cube algorithm. 
So SDF or NeRF has an advantage in tasks like inverse rendering that rely on a good surface representation. 
However, due to the dense sampling in 3D space, its rendering is less efficient and limits its applicability on consumer-level devices. 
Moreover, SDF and NeRF are less successful in dynamic scene reconstruction due to their implicit representation. 
}

\wtmin{
\textbf{\shortgs.}
\shortgs also has explicit geometry but different from mesh it has no edges or faces that connect different Gaussians. 
To make up for the missing connection information, each Gaussian has anisotropic scales to fill the gap between neighboring Gaussians and synthesize realistic novel views. 
With the rasterization-based renderer, \shortgs allows real-time visualization of 3D scenes on consumer-level devices, which makes applications like large-scene reconstruction, SLAM, and generation that have efficiency requirements more applicable. 
Apart from the efficient rendering, its explicit geometry representation enables flexible point reprojection from one viewpoint to another viewpoint, making simultaneous geometry reconstruction and camera pose optimization and dynamic reconstrution easier and tasks like SLAM and large scene reconstruction more efficient.
However, due to the discrete geometry representation, the geometry reconstruction quality of current \shortgs-based methods is only comparable to previous SDF-based methods like NeuS~\cite{NeuS}. 
It would be promising to combine other 3D representations with the \shortgs similar to~\cite{gao2024mesh,NeuSG} to build up high-quality geometry/surface, facilitating downstream applications like automated vehicles and animations.
}

After discussing the advantages and disadvantages of different 3D representations, we summarize the challenges that remain for for the \longgs and how these challenges might be resolved in the future.

\textbf{Robust and generalizable novel view synthesis.} 
Although \longgs has achieved realistic novel view synthesis results, its reconstruction quality degrades as indicated by~\cite{StopThePop} when dealing with challenging inputs like sparse-view inputs, complex shading effects, and large-scale scenes. 
Though attempts~\cite{SparseGS,GaussianShader,DrivingGaussian} have been made to get better results, there are still improvement spaces. 
How to improve its reconstruction robustness on different inputs is an important problem. 
\wtmin{In addition, developing a generalizable reconstruction pipeline with or without data prior like~\cite{pixelSplat,SplatterImage,xu2024agg} would significantly reduce the training cost.}

\textbf{Geometry reconstruction.}
Despite the efforts on the rendering quality side, few methods~\cite{NeuSG,SuGaR} work on geometry/surface reconstruction with the \shortgs representation. 
Compared to the continuous implicit representations like NeRF and SDF, \shortgs's geometry quality still suffers from its discrete geometry representation.

\textbf{Independent and efficient 3D editing.}
A few methods have dived into the field of editing \longgs's geometry~\cite{gao2024mesh,huang2023sc,waczynska2024games,SuGaR}, texture~\cite{GaussianEditor,GaussianEditor_text}, and lighting~\cite{GaussianShader,GS-IR,RelightableGaussian,GIR}. 
However, they cannot decompose geometry, texture, and lighting accurately or need re-optimization of Gaussians' attributes. 
As a result, these methods still miss independent editing capabilities or lack efficiency in the editing process. 
It is promising to extract geometry, texture, and lighting with more advanced rendering techniques to facilitate independent editing and build the connection between \shortgs and mesh-based representation to enable efficient editing. 

\textbf{Realistic 4D generation.}
With the help of SDS loss based on SD~\cite{Diffusion}, generative models~\cite{zou2023triplane,tang2024lgm,tang2023dreamgaussian} with the \shortgs representation have produced faithful results. 
However, 4D generation results by current methods~\cite{blattmann2023align,ren2023dreamgaussian4d,yin20234dgen} still miss realistic geometry, appearance, and physics-aware motion. 
Integrating data prior like results produced by video generative models and physical laws might boost the quality of generated 4D content. 

\textbf{Platform.} 
Most implementations of methods and frameworks like GauStudio~\cite{GauStudio} for the \longgs representation are written in Python with the cuda-supported PyTorch~\cite{PyTorch} framework, which may limit its future applicability on wider platforms. 
Reproducing it with deep learning frameworks like Tensorflow~\cite{TensorFlow} and Jittor~\cite{Jittor} can facilitate its usage on other platforms.

\appendix

\subsection*{Declarations}
\paragraph{Availability of data and materials}
As the paper does not involve the generation or analysis of specific datasets, there are no associated data or materials. 

\paragraph{Competing interests}
The authors have no competing interests to declare that are relevant to the
content of this article.

\paragraph{Funding}
This work was supported by National Natural Science Foundation of China (No. 62322210), Beijing Municipal Natural Science Foundation for Distinguished Young Scholars (No. JQ21013), and Beijing Municipal Science and Technology Commission (No. Z231100005923031).

\paragraph{Authors' contributions}
Tong Wu conducted an extensive literature review and drafted the manuscript. 
Yu-Jie Yuan, Ling-Xiao Zhang, and Jie-Yang provided critical insights, analysis of the existing research, and part of the manuscript writing. 
Yan-Pei Cao, Ling-Qi Yan, and Lin Gao conceived the idea and scope of the survey and improved the writing. 
All authors read and approved the final manuscript.

\paragraph{Acknowledgements}
We would like to thank Jia-Mu Sun and Shu-Yu Chen for their suggestion in the timeline figure.

\bibliographystyle{CVMbib}
\bibliography{main}

\subsection*{Author biography}

\begin{biography}[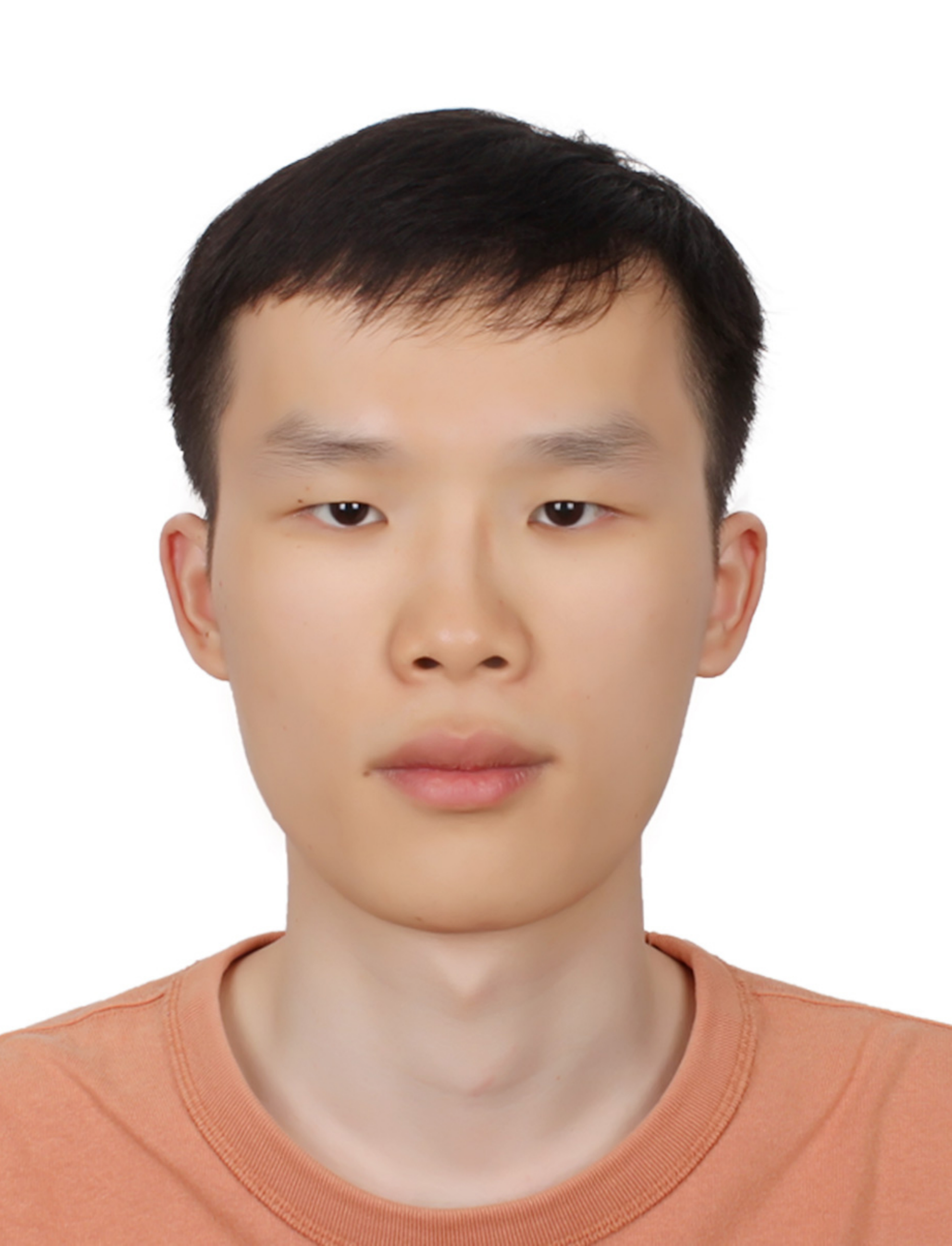]{Tong Wu} received his bachelor’s degree in computer science from Huazhong University of Science and Technology in 2019. He is currently a PhD candidate at the Institute of Computing Technology, Chinese Academy of Sciences. His research interests include computer graphics and computer vision. 
\end{biography}

\begin{biography}[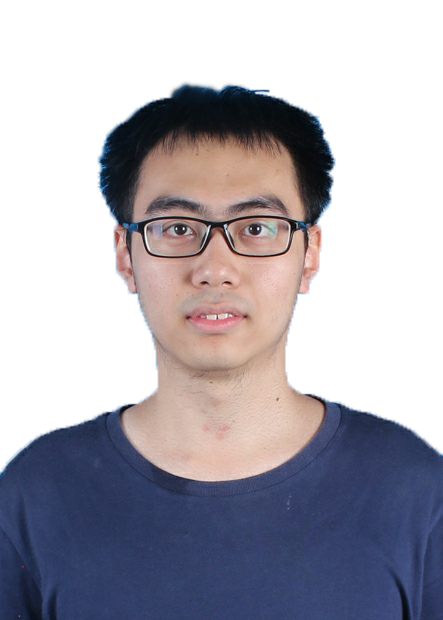]{Yu-Jie Yuan} received the bachelor’s degree in mathematics from Xi'an Jiaotong University in 2018. He is currently a Ph.D. candidate at the Institute of Computing Technology, Chinese Academy of Sciences. His research interests include computer graphics and neural rendering.
\end{biography}

\begin{biography}[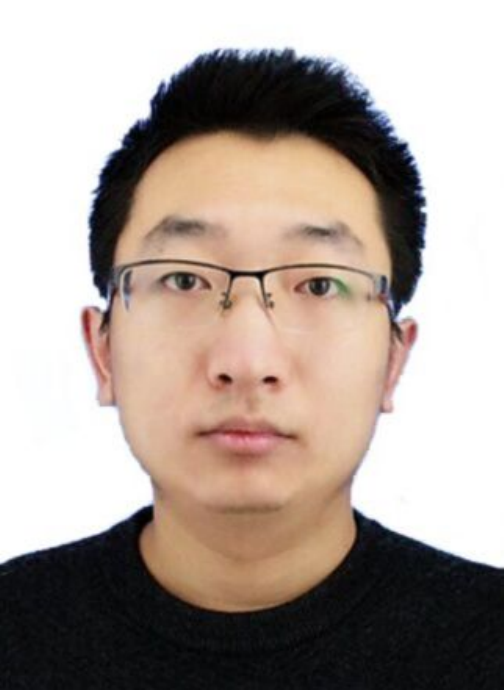]{Ling-Xiao Zhang} received the master of engineering's degree in computer technology from the Chinese Academy of Sciences in 2020. He is currently an engineer at the Institute of Computing Technology, Chinese Academy of Sciences. His research interests include computer graphics and geometric processing.
\end{biography}

\begin{biography}[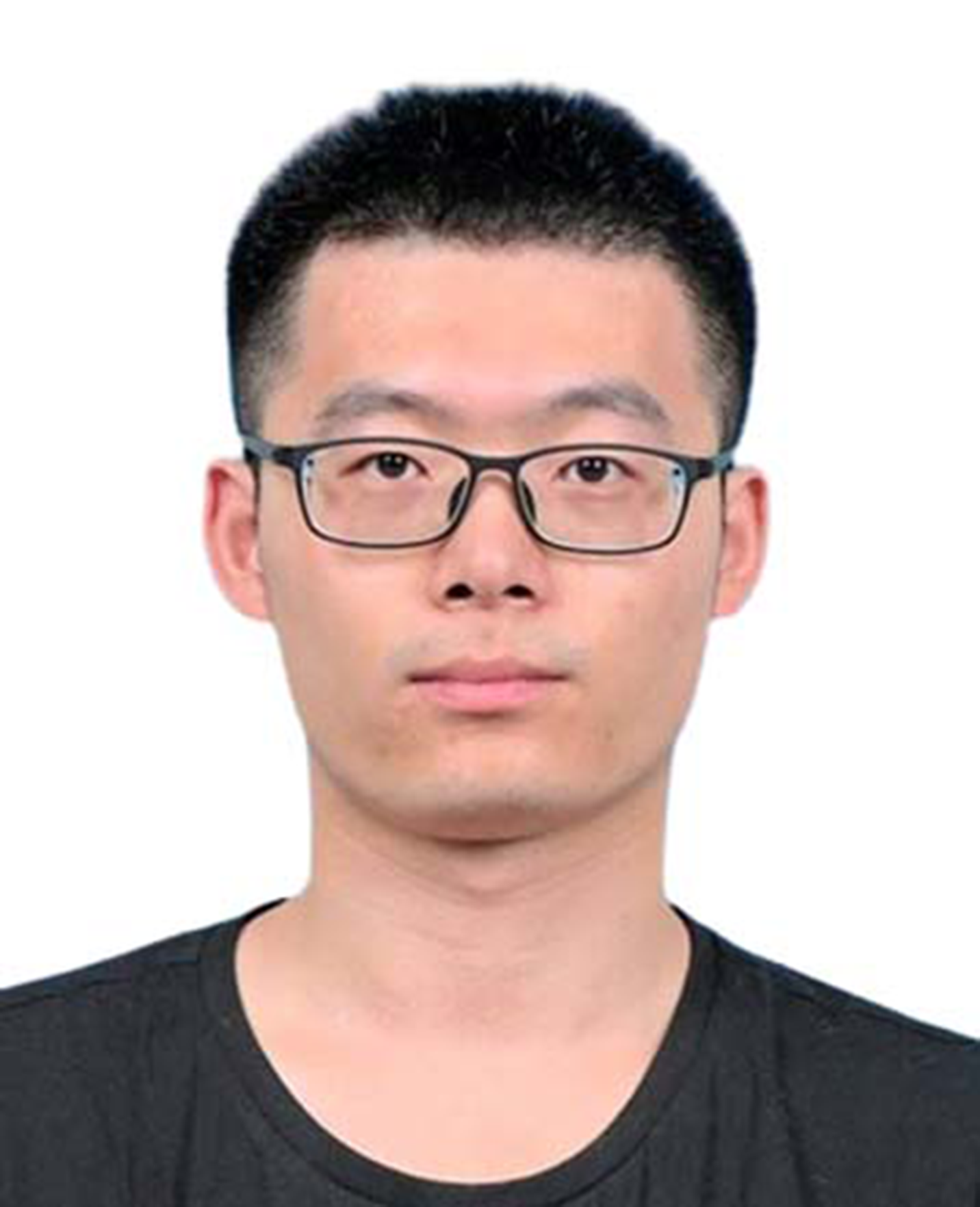]{Jie Yang} received a bachelor's degree in mathematics from Sichuan University and a Ph.D. degree in computer science from Institute of Computing Technology, Chinese Academy of Sciences. He is currently an Assistant Professor at the Institute of Computing Technology, Chinese Academy of Sciences. His research interests include computer graphics and geometric processing.
\end{biography}

\begin{biography}[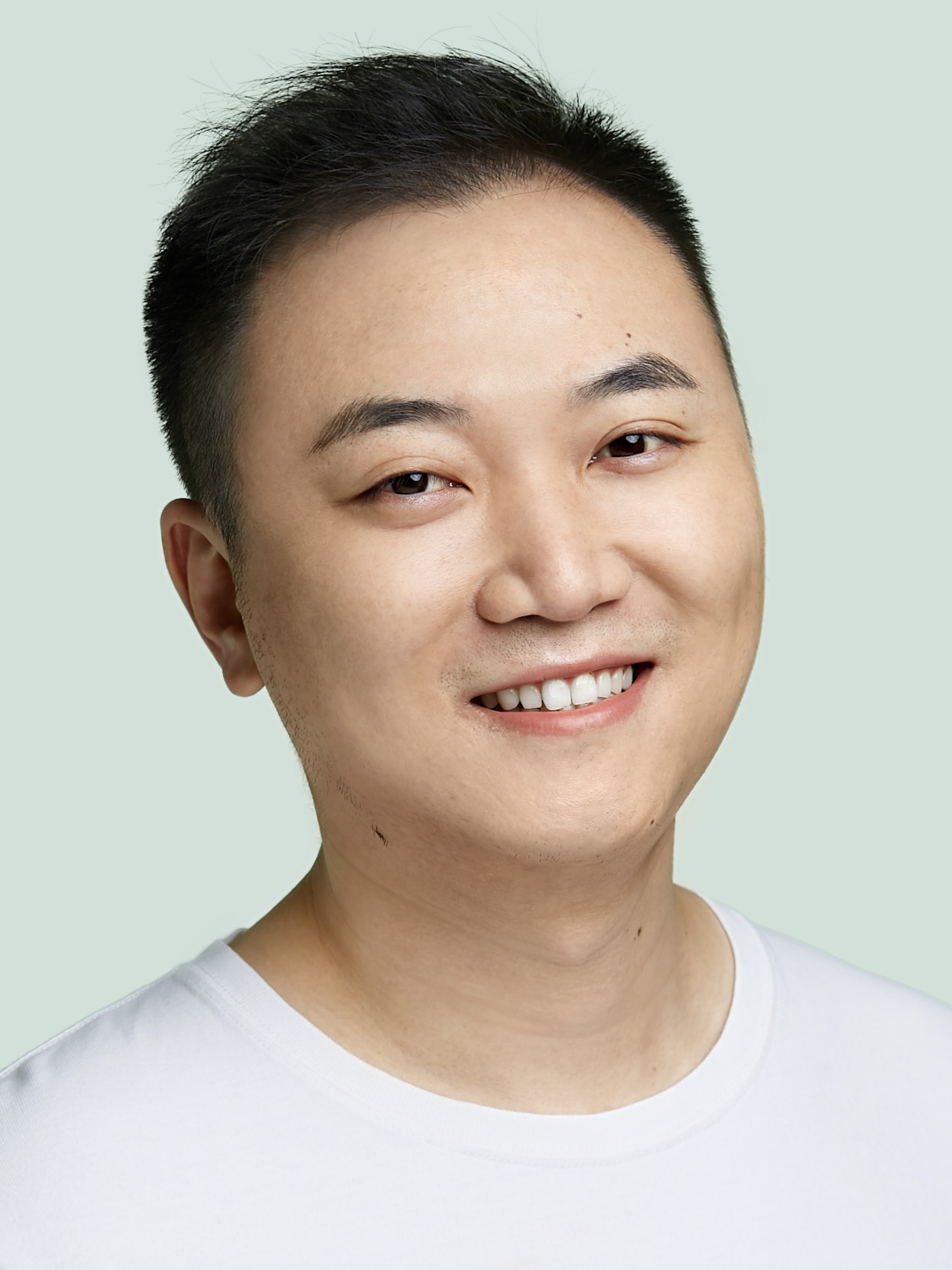]{Yan-Pei Cao} received bachelor’s and Ph.D. degrees in computer science from Tsinghua University in 2013 and 2018, respectively. 
He is currently the Head of Research and Founding Team at VAST. His research interests include computer graphics and 3D computer vision.
\end{biography}

\begin{biography}[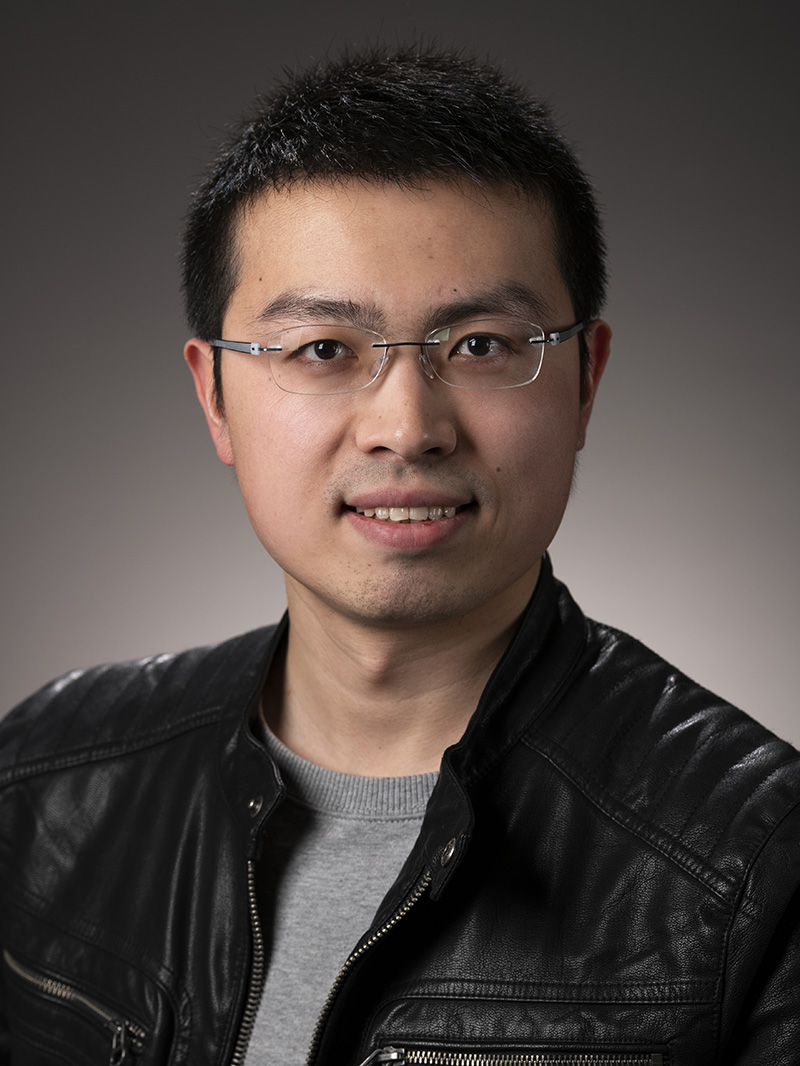]{Ling-Qi Yan} is an Assistant Professor of Computer Science at UC Santa Barbara, co-director of the MIRAGE Lab, and affiliated faculty in the Four Eyes Lab. Before joining UCSB, he received his Ph.D. degree from the Department of Electrical Engineering and Computer Sciences at UC Berkeley. 
\end{biography}

\begin{biography}[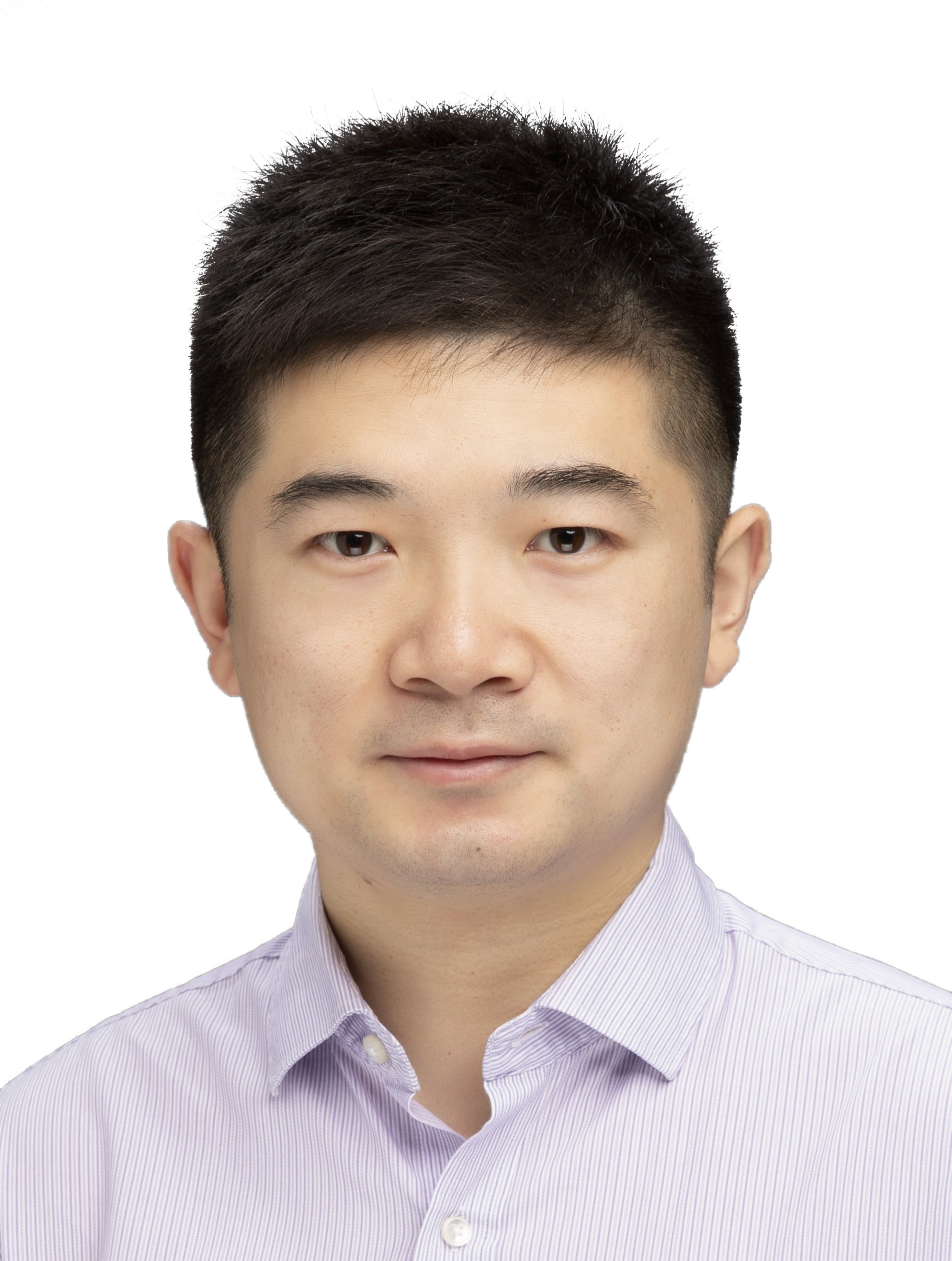]{Lin Gao} received the bachelor’s degree in mathematics from Sichuan University and the PhD degree in computer science from Tsinghua University. He is currently a Professor at the Institute of Computing Technology, Chinese Academy of Sciences. He has been awarded the Newton Advanced Fellowship from the Royal Society and the Asia Graphics Association young researcher award. His research interests include computer graphics and geometric processing.
\end{biography}

\end{document}